\pdfoutput=1

\documentclass[11pt]{article}

\usepackage[]{acl}
\usepackage{times}
\usepackage{latexsym}
\usepackage{colortbl}
\usepackage{graphics}
\usepackage{graphicx}
\usepackage{multirow}
\usepackage{amssymb}
\usepackage{mathrsfs}
\usepackage{amsmath}
\usepackage{bm}
\usepackage[cmintegrals]{newtxmath}
\usepackage{subfigure}
\usepackage{caption}
\usepackage{makecell}
\usepackage{threeparttable}
\usepackage{booktabs}
\usepackage{tabularx}
\usepackage{array}
\usepackage{makecell}
\usepackage{booktabs}
\usepackage{pifont}
\usepackage{bigstrut}

\newcommand{\augloss}{$\mathcal{L}_{\textit{aug}}$}
\newcommand{\repreloss}{$\mathcal{L}_{\textit{repre}}$}
\newcommand{\logitsloss}{$\mathcal{L}_{\textit{logits}}$}
\def \redcell#1 {\cellcolor[HTML]{FDDCCE}{#1}}
\def \greencell#1{\cellcolor[HTML]{AFF7C2}{#1}}
\def \Greencell#1{\cellcolor[HTML]{72EB70}{#1}}
\newcommand{\ro}{$\rho$}

\usepackage[T1]{fontenc}

\usepackage[utf8]{inputenc}

\usepackage{microtype}
\usepackage{inconsolata}
\title{Towards Robust and Generalizable Training: An Empirical Study of Noisy Slot Filling for Input Perturbations}

\author{Jiachi Liu $^{1}$\thanks{\ \ The first three authors contribute equally. Weiran Xu is the corresponding author. Email: dongguanting@bupt.edu.cn},
Liwen Wang$^{1*}$, 
Guanting Dong$^{1*}$,
Xiaoshuai Song$^{1}$,
Zechen Wang$^{1}$,
\\
\textbf{Zhengyang Wang$^{1}$,
Shanglin Lei$^{2}$,
Jinzheng Zhao$^{4}$,
Keqing He$^{3}$,
Bo Xiao$^{1}$,
Weiran Xu$^{1}$}    \\
$^{1}$Beijing University of Posts and Telecommunications, Beijing, China\\
$^{2}$Huazhong University of Science and Technology , Wuhan , China\\
$^{3}$Meituan Group, Beijing, China\\
$^{4}$University of Surrey, UK\\
}

\begin{document}
\maketitle
\begin{abstract}
In real dialogue scenarios, as there are unknown input noises in the utterances, existing supervised slot filling models often perform poorly in practical applications. Even though there are some studies on noise-robust models, these works are only evaluated on rule-based synthetic datasets, which is limiting, making it difficult to promote the research of noise-robust methods. In this paper, we introduce a noise robustness evaluation dataset named Noise-SF for slot filling task. The proposed dataset contains five types of human-annotated noise, and all those noises are exactly existed in real extensive robust-training methods of slot filling into the proposed framework. By conducting exhaustive empirical evaluation experiments on Noise-SF, we find that baseline models have poor performance in robustness evaluation, and the proposed framework can effectively improve the robustness of models. Based on the empirical experimental results, we make some forward-looking suggestions to fuel the research in this direction. Our dataset Noise-SF will be released at https://github.com/dongguanting/Noise-SF.



\end{abstract}

\section{Introduction}

The slot filling (SF) task in Spoken Language Understanding (SLU) aims to automatically extract the informative slot values from speakers' utterances. Recent data-driven supervised learning methods, based on sequence labeling framework, have been applied to slot filling and achieved excellent results \cite{liu-lane-2016-joint,Liu2016AttentionBasedRN,goo-etal-2018-slot,he-etal-2020-syntactic,he-etal-2020-learning-tag,yan-etal-2020-adversarial,wang-etal-2021-bridge,zhao-etal-2022-entity,10095149,dong2023multi,li-etal-2023-generative}. However, the high performance of these methods depends heavily on the distribution consistency between training data and test data. As the data distribution of dialogues in real scenarios is unknown \cite{wu2021bridging}, there are still many challenges in applying these methods to real dialogue scenarios.

\begin{figure}[t]
\centering
\resizebox{.44\textwidth}{!}{\includegraphics{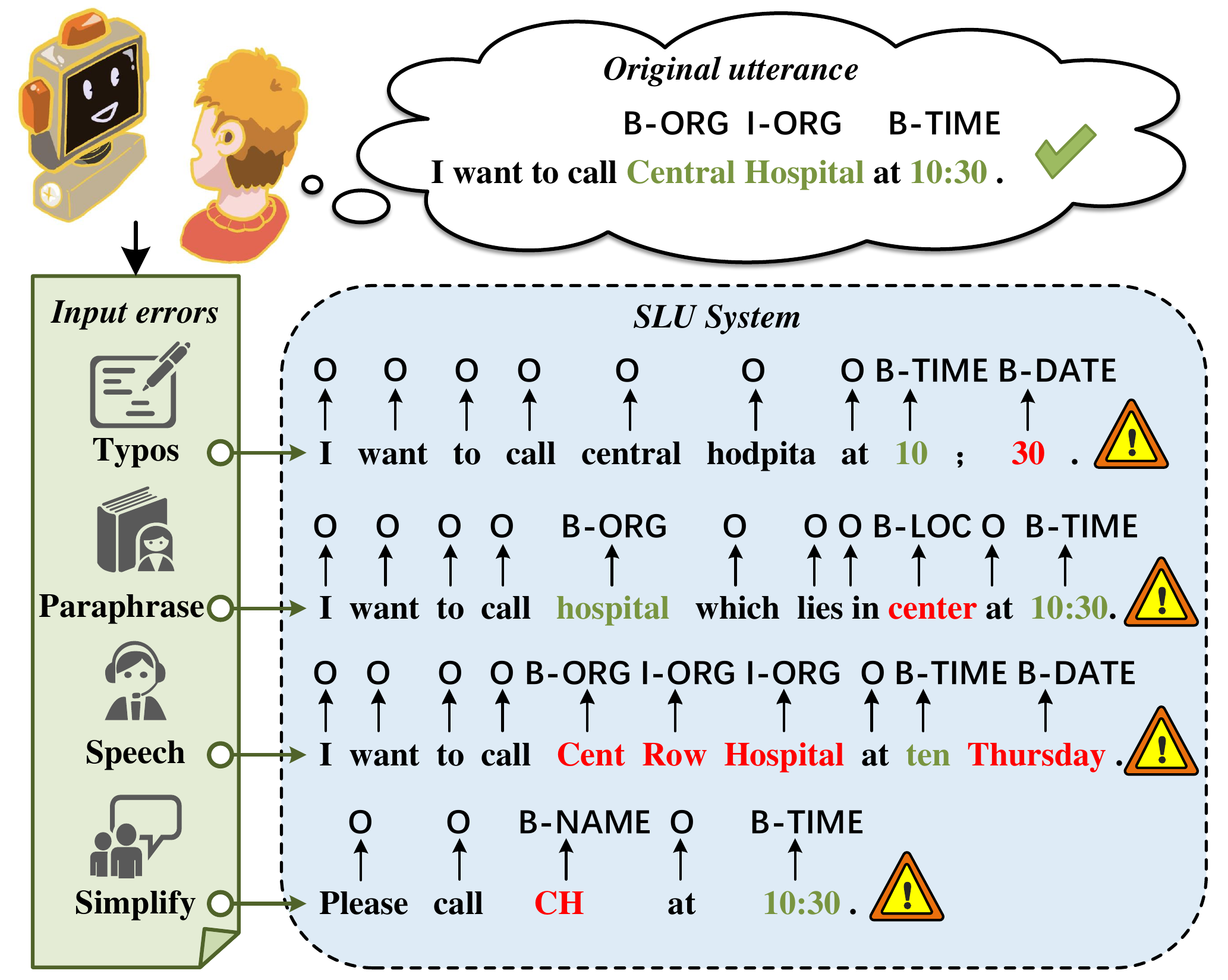}}
\vspace{-0.2cm} 
\caption{The impact of various types of input noise on the slot filling\label{fig:intro2} system in real scenarios}

\vspace{-0.2cm} 
\end{figure}
\begin{table}[htbp]
\centering
\scriptsize
\setlength\tabcolsep{1.6pt}
\renewcommand\arraystretch{1.5} 
\begin{tabular}{|c|c|ccccc|} 
\hline
& Clean & Typos & Speech & Paraphrase & Verbosity & Simplification  \\
\hline
F1 & 95.8 & 64.3 (\textcolor[RGB]{200,42,42}{\textbf{-31.5}}) & 82.1 (\textcolor[RGB]{200,42,42}{\textbf{-13.7}}) & 88.1 (\textcolor[RGB]{200,42,42}{\textbf{-7.7}}) & 82.2 (\textcolor[RGB]{200,42,42}{\textbf{-13.6}}) & 85.9 (\textcolor[RGB]{200,42,42}{\textbf{-9.9}})\\
\hline
\end{tabular}
\caption{Performance drop of baselines on Noise-SF}
\label{tbl:intro1}
\vspace{-0.5cm} 
\end{table}

As shown in Table \ref{tbl:intro1}, the slot filling model trained on the noise-free dataset has a significant performance degradation when facing the noisy input. Due to the abundant language expression and input errors, there are various types of input noise in practical applications. 
As illustrated in Figure \ref{fig:intro2}, when expressing the original intents, users may not interact with the dialogue system following the standard input format due to the difference in expression habits. Instead, they may paraphrase or simplify their utterances. In addition, the slot filling model also shows poor performance due to errors from the upstream input system. For instance, there are typos from the keyboard input, and speech errors caused by the ASR systems. Therefore, it is necessary to train a robust slot filling model against the input noise, which facilitates the task-oriented dialogue system to be widely applied in real scenarios \cite{zeng2022semi}.

Recently, some works have focused on the impact of noisy input text on Natural Language Processing (NLP) systems.  Most existing studies \cite{wu2021bridging,moradi2021evaluating,gui2021textflint} explore the robustness problem on rule-based synthetic datasets, which have certain limitations. 
To explore the robustness of the SLU system against ASR errors,
\citet{huang2020learning,ruan2020towards,gopalakrishnan2020neural,li2020multi} use aligned human-transcribed texts to derive a noisy slot filling dataset. 
\citet{peng2020raddle} propose RADDLE, an evaluation benchmark for robust Dialog State Tracking (DST). Based on MultiWOZ \cite{budzianowski2020multiwoz}, they annotate various noisy real task-oriented dialogue utterances to evaluate the end-to-end dialogue system and the DST task. Unfortunately, due to the lack of text-label pairs with real noises for slot filling tasks, it's hard to promote the research of noise-robust slot filling methods.

To address the above challenges, we intend to explore the effects of different robust-training strategies toward various input noise that exist in real dialogue scenarios, and provide empirical guidance for the research of robust slot filling methods. 
In this paper, we first extract the utterances with human-annotated noise from RADDLE and construct a slot filling dataset named Noise-SF through manual annotation. Compared to the rule-based synthetic robustness evaluation dataset, the proposed dataset contains five types of noise existed in realistic scenarios, and it is more natural. To explore the effects of different robust-training strategies, we propose a universal noise-robust training framework for slot filling models. 
The proposed framework includes five text-level and four feature-level augmentation methods, and we can apply it to generate robust samples and introduce extra consistency training objectives.
Based on this framework, we conduct extensive experiments on Noise-SF and analyze the results from the perspectives of noise types and robust training methods, respectively. Besides, we also design some combination experiments to explore the best combination modes of feature-level and text-level data augmentation methods. And we further construct a multi-noise mixed dataset to verify the noise-robust effect in mixed noise scenarios. 


Our contributions are four-fold: 
1) To fuel the research of robust slot filling methods, we introduce a slot filling dataset containing five types of real input noise; 
2) We propose a universal noise-robust training framework which includes five text-level and four feature-level augmentation methods for slot filling tasks; 
3) We conduct extensive comparison and combination experiments on Noise-SF based on the proposed training framework, and comprehensive analyses from the perspectives of noise type and robust training methods, respectively; 
4) From the perspectives of slot entity mentions and contextual semantics, we conclude the experimental phenomena and provide some constructive advice for further research.

\section{Related Work}
\paragraph{Noise-input Problem}

As there is still a big performance gap between the real-world scenarios and benchmarks \cite{belinkov2017synthetic,check2020}, recently, the robustness of NLP systems against input perturbations has attracted a lot of attention. \newcite{eval2021} present empirical evaluations of the robustness of various NLP systems against input perturbations on synthetic generated benchmarks. \newcite{namysl2020nat, nat2021} focus on the robustness of the NER model against Optical Character Recognition (OCR) noise and misspellings. Compared to other NLP systems, dialogue systems would face more diverse input noise due to more frequent interactions with users. \newcite{fang2020using,gopalakrishnan2020neural} investigate the robustness of dialogue systems on ASR noise, and 
\newcite{ruan2020towards,li2020improving,huang2020learning,li2020multi,wu2021bridging} mainly focus on the ASR-noise-robustness SLU models in dialogue systems. Although it is crucial to build robust NLP systems, there are few benchmarks available for deeper investigations. Most of existing studies simulate the real noise with rule-based methods \cite{wu2021bridging,moradi2021evaluating,gui2021textflint,liu2020robustness,dong2022pssat,10094766}, which have certain limitations. 
To further explore this direction, 
RADDLE  \cite{peng2020raddle} offers a crowd-sourced robustness evaluation benchmark for dialog systems, which includes various noisy utterances existed in real dialogue scenarios. 

In this paper, we investigate the robustness on various types of real noise for slot filling models, and conduct exhaustive comparison and experiments for further research in this direction.

\paragraph{Robust Training Methods}
Noisy data augmentation methods are widely applied to generate robust samples for noise-robust training.
\newcite{heigold-etal-2018-robust} use word scrambling, character flips and swaps to augment training data. \newcite{belinkov2017synthetic} use possible lexical replacements from Wikipedia edit histories as a natural source of the noise. \newcite{namysl2020nat} use a character confusion matrix to model real-world noise. These augmentation methods only introduce noise at the text-level. To explore higher-dimensional data augmentation methods, \newcite{yan2021consert} explore the effectiveness of four feature-level data augmentation methods. 
\newcite{lei2023watch,lei2023instructerc} presents several sub tasks and explore sentence-level noise disturibance to avoid the model overfitting to a specific dataset. \newcite{dong2022pssat,10094766} introduce external unsupervised knowledge base to help model capture noisy semantic structure. However, this undoubtedly introduces a lot of additional computing resources and noise.


\section{Problem Formulation}
\subsection{Slot Filling Task}
The slot filling task is usually formulated as a sequence labeling problem. Given a tokenized input sentence $ X = \{ x_{1}, ..., x_{N}\} $ of length $ N$, the slot filling task aims to predict a corresponding tag sequence $ Y= \{ y_{1}, ..., y_{N}\} $ in BIO format, where each $ y_{n}$ can take three types of values: $ B\_slot\_type $, $ I\_slot\_type $, and O, standing for the \textbf{B}eginning, \textbf{I}ntermediate word of a slot and the word \textbf{O}utside slot, respectively. Existing models for slot filling task employ Bi-LSTM \cite{lample2016neural} or pre-trained language models \cite{namysl2020nat} as encoder and a softmax layer \cite{chiu2016named} (with a Conditional Random Field \cite{lample2016neural}) as decoder.
\vspace{-0.2cm}
\subsection{Noise Input Robust Slot Filling}

We use $D_{train}, D_{dev}$ and $D_{test}$ to represent the conventional labeled slot filling dataset, where $D_{test} = \{X_1:Y_1, X_2:Y_2, ..., X_T :Y_T\}$ of length $T$. We define $\mathcal{P}_i=X_{i}:Y_{i}$ and then $D_{test}=\{\mathcal{P}_1,\mathcal{P}_2,\dots,\mathcal{P}_T\}$. Given $D_{train}, D_{dev}, D_{test}$ and a real unknown noise process $\Gamma = P(\widetilde{X}: \widetilde{Y} | X: Y )$, $\Gamma$ can be regarded as the transformation of adding real noise to the input text and corresponding labels in the slot filling system \cite{brill2000improved}. The noise-input-robust test set is defined in (1):
\vspace{-0.1cm}
\begin{equation}
\begin{aligned}
\widetilde{\mathcal{P}}_i = \widetilde{X}_i : \widetilde{Y}_i = \Gamma(X_i,Y_i)\ \\ 
\widetilde{D}_{test} = \{\widetilde{\mathcal{P}}_1,\widetilde{\mathcal{P}}_2,\dots,\widetilde{\mathcal{P}}_T \}
\end{aligned}
\end{equation}
In real scenarios, the form of the noise transformation $\Gamma$ is flexible, which could be a specific type of noise (e.g., Typos) or mixed noise.


We introduce the noise-input slot filling task, which is defined as training models on a clean training set and then testing on the $\widetilde{D}_{test}$ with input noises. The goal is to make the model achieve the best performance on the noise dataset $\widetilde{D}_{test}$.
\vspace{-0.2cm}

\section{Dataset}
To our knowledge, there are no existing evaluation datasets including natural noisy slot filling data pairs. To explore the noise-input slot filling problem, in this section, we introduce a new dataset named Noise-SF.
\vspace{-0.2cm}
\subsection{Data Collection}
We build Noise-SF based on RADDLE \cite{peng2020raddle} and MultiWOZ-SF \cite{lu-etal-2021-slot}. RADDLE is a crowdsourced diagnostic evaluation dataset covering a broad range of real-world noisy texts for end-to-end dialog systems and dialog state tracking. In particular, the original dataset of the evaluation set in RADDLE is extracted from MultiWOZ \cite{budzianowski2020multiwoz}. The detail of data processing is in Appendix \ref{appendixa}.

\subsection{Data Construction}
For MultiWOZ-SF, it uses a total number of 61119 clean utterances including four domains (attraction, hotel, restaurant, and train) for training, while randomly selecting 5000 utterances to form the validation set. We extract clean utterances and five kinds of noisy utterances (Typos, Speech, Simplification, Verbosity, and Paraphrase) from RADDLE when constructing the test set. \textbf{Paraphrase} widely exists in different users who use other words to restate the text following their language habits, \textbf{Verbosity} refers to users using redundant words to express intentions \textbf{Simplification} refers to users using concise words to express intentions, \textbf{Typos} is often caused by non-compliant abbreviations; \textbf{Speech} is caused by recognition and synthesis errors 
from ASR systems. We categorize Typos to character-level noise, speech to word-level noise, and Simplification, Verbosity, Paraphrase to sentence-level noise. The Statistic of Noise-SF are in Appendix \ref{appendixb}, and the consideration for noise evaluation set is in Appendix \ref{appendixc}.

\begin{figure*}[ht]
\centering
\resizebox{.8\textwidth}{!}{\includegraphics{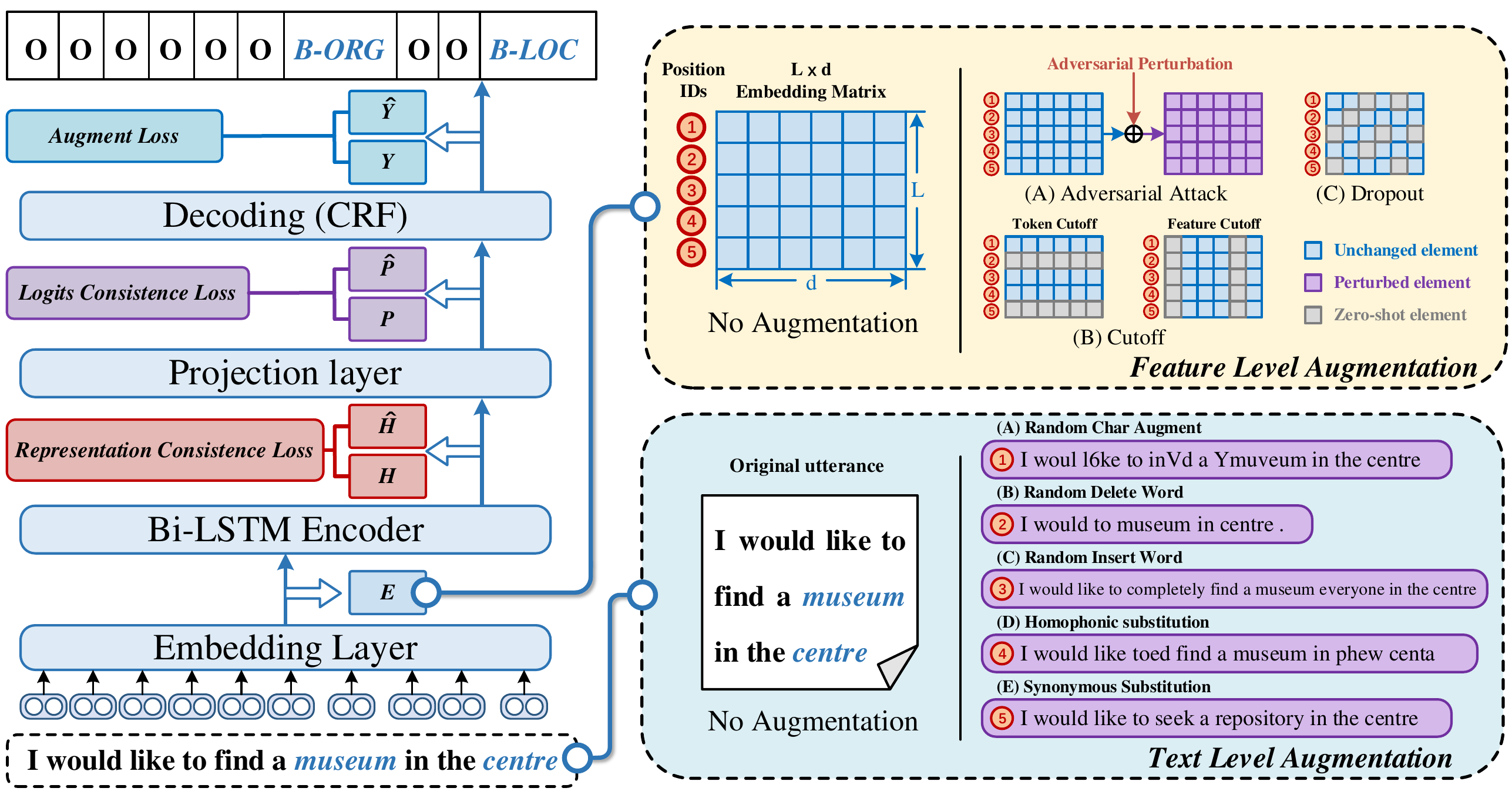}}
\vspace{-0.3cm}
    \caption{The overall architecture of the noise-robust training framework (left) and the two-level data augmentation strategies (right).}
 
\label{fig:method}
\vspace{-0.6cm} 
\end{figure*}

\vspace{-0.2cm}
\subsection{Evaluation Metrics}
To evaluate the performance of baselines and noise-robust training methods in the task of slot filling with noisy input, it is necessary to define metrics that can reflect the impact of noise and the performance of robustness methods. First, we introduce $\Delta F1=F1^{clean}-F1^{noise}$ to calculate the difference of F1 scores on the noise-input evaluation dataset compared to the clean text set. $\Delta F1$ reflects the extent to which different models are affected by different types of noise. Secondly, we introduce $\mathcal{R}_{NRM}= \Delta F1_{NRM}-\Delta F1_{baseline}$, where $\Delta F1_{NRM}$ indicates the $\Delta F_{1}$ of noise-robust methods, and $\Delta F1_{baseline}$ indicates the $\Delta F1$ of baselines. The values of $\mathcal{R}_{NRM}$ reflect noise-input robustness of different noise-robust methods. However, as illustrated in Table \ref{tbl:intro1}, different noises show different impacts on the baseline model, where the F1 of the baseline on the Typos noise is 31.52. Therefore, the noise-robust model could have a larger performance improvement on this type of noise than others. To better reflect the noise-robustness performance of noise-robust methods on different types of noise, we introduce the relative denoising rate $\rho=\mathcal{R}/\Delta F1_{baseline}$. By comparing the $\rho$ value of the input robustness method on different noise types, it can reflect which noise the method is more effective for.
\vspace{-0.2cm}

\section{Methodology}
\vspace{-0.2cm}
This section introduces our noise-robust training framework. The overall structure of the proposed framework is shown in Figure \ref{fig:method}.
\vspace{-0.2cm}

\subsection{Noise-robust Training}
The noise-robust model consists of a backbone network of Embedder-Encoder-CRF structure and three consistency training strategies. Given an input utterance $X=\{x_1,x_2,\dots,x_n\}$, we get $E=\{e_1,e_2,\dots,e_n\}$ after embedder. Then, the embeddings of all tokens are forward into the BiLSTM-based encoder to get the hidden representation of the sentence input $H=\{h_1,h_2,\dots,h_n\}$. In the next step, the hidden features are passing through the projection layer to get the classification logits $P=\{p_1,p_2,\dots,p_n\}$. In the end, these logits are feed into CRF layer and get the final labels of each token $Y=\{y_1,y_2,\dots,y_n\}$. The normal training object function of slot filling model is $\mathcal{L}_{normal}=\text{CrossEntropy}(\tilde X, \tilde Y)$

To ensure the consistency of the output results when the input is disturbed, inspired by a series augmentation works on loss level \cite{chakraborty2018adversarial,9747192,dong2023bridging,10193387}, we propose three consistency training strategies to constrain the consistency of the representation of the original sample X and the augmented sample $\tilde X$ at each layer.

\textbf{Representation Consistency Loss} (\repreloss) We calculate the MSELoss between the original sample representation $H$ and the augmented sample representation $\tilde H$ in the hidden layer as the representation consistency loss $\mathcal{L}_{consis}={\rm MSELoss}(H, \tilde H)$. The model will constrain the stability of the representation in the hidden layer by optimizing $\mathcal{L}_{repre}$.  

\textbf{Logits Consistency Loss} (\logitsloss) We calculate Kullback–Leibler divergence between the classification logits of the original sample and the logits of the augmented sample to get the logits consistency loss $\mathcal{L}_{consis}=KL{\rm -Divergence}(P, \tilde P)$. The optimization object of $L_{logits}$ is to enable the model to maintain consistency of the logits in the classification layer when disturbance appears.

\textbf{Augmentation Loss} (\augloss) We directly calculate the cross-entropy loss of the augmented sample and the corresponding label: $\mathcal{L}_{consis}=\text{CrossEntropy}(\tilde X, \tilde Y)$. Then, we obtain the final objective $\mathcal{L} =\mathcal{L}_{normal}+\alpha \cdot \mathcal{L}_{consis}$

\vspace{-0.2cm}
\subsection{Text-level Augmentation}

To model the input noise existing in realistic scenarios, we introduce five well-designed Text-level augmentation methods:

\textbf{Random Char Augmentation} (\emph{CharAug}) is a character-level augmentation method which randomly adds, deletes and replaces characters in one token with the transform probability $p$. And our settings are consistent with the construction principle of NAT\footnote{The training objective functions $\mathcal{L}_{augm}$ and $\mathcal{L}_{stabil}$ in NAT are corresponded to $\mathcal{L}_{aug}$ and $\mathcal{L}_{logits}$ in our works, respectively} \cite{namysl-etal-2020-nat}.
\textbf{Random Word Deletion} (\emph{DeleteWord}) aims to model the impact of Simplification noise on input utterance in real scenarios \cite{wei2019eda}, which randomly deletes tokens with probability $p$.
\textbf{Random Word Insertion} (\emph{InsertWord}) randomly inserts words with probability $p$ based on contextual embedding \cite{peng2020data}.
\textbf{Homophonic substitution} (\emph{SpeechAug}) is designed for modeling Speech errors. We implement a homophone replacement dictionary, where words in the utterance are replaced with homophones with probability $p$.
\textbf{Synonymous Substitution} (\emph{SubWord}) is implemented based on WordNet's \cite{miller1995wordnet} synonymous thesaurus. We randomly select tokens in utterance with probability $p$ for synonymous substitution \cite{coulombe2018text}. 


\vspace{-0.3cm}
\subsection{Feature-level Augmentation}
\vspace{-0.1cm}

In this paper, we apply four different feature-level data augmentation strategies to augment the representation $e(x)$. Inspired by \newcite{yan2021consert}, these strategies are introduced to inject perturbation into the model while training:

\textbf{Adversarial Attack} (Adv) is generally applied to improve the model’s robustness \cite{goodfellow2014explaining}. We implement this strategy with Fast Gradient Value (FGV) \cite{rozsa2016adversarial}, which directly uses the gradient to compute the perturbation. 
\textbf{TokenCut} is a simple and efficient data augmentation strategy proposed by \newcite{shen2020simple}. We randomly erase some tokens in the $L \times d$ feature matrix. 
\textbf{FeatureCut} is also proposed by \newcite{shen2020simple}. We randomly erase some feature dimensions in the $L \times d$ feature matrix. 
\textbf{Dropout} is a widely used regularization method to avoid overfitting \cite{hinton2012improving}. We randomly drop elements in the token embedding layer by a pre-defined probability and set the values to zero. Note that this strategy is different from Cutoff since each element is considered individually. 
\vspace{-0.2cm}
\section{Experiments}
\vspace{-0.1cm}
\begin{table*}[t!]
    \centering
    \tiny
    \setlength{\tabcolsep}{3pt}
    \renewcommand{\arraystretch}{1.5}
    \begin{tabular}{c | c | c | c| c |c |c c c c |c}
    \toprule[1pt]
\multirow{2}{*}{\textbf{Model}}  & \multirow{2}{*}{\textbf{Augmentation type}} & \multirow{2}{*}{\textbf{Loss type}}
& \multirow{2}{*}{\textbf{Clean test\ding{172}}} & \textbf{Character-level}  &  \textbf{Word-level} 
& \multicolumn{4}{c}{\textbf{Sentence-level}} \vline & \multirow{2}{*}{\textbf{Overall\ding{179}}} \\

\cline{5-10}
 & & & & \textbf{Typos\ding{173}} & \textbf{Speech\ding{174}} & \textbf{Paraphrase\ding{175}} 
& \textbf{Verbosity\ding{176}} & \textbf{Simplification\ding{177}} & \textbf{Overall\ding{178}} & \\

\hline

Original & none & $\mathcal{L}$
& \underline{95.8} & \makecell{\underline{64.3 (-31.5)}} & \makecell{\underline{82.1 (-13.7)}} & \makecell{\underline{88.1 (-7.7)}} 
& \makecell{\underline{82.2 (-13.6)}} & \makecell{\underline{85.9 (-9.9)}} & \underline{85.4 (-10.4)} &  \makecell{\underline{80.5 (-15.3)}}\\

\hline
\multirow{9}{*}{\makecell{Text-level}} & \multirow{3}{*}{CharAug (NAT)} & \augloss &
96.0 (0.2) & \textbf{\Greencell{76.3 (38.1\%)}} & 84.1 (14.6\%) & \cellcolor[HTML]{FDDCCE}{87.8 (-3.9\%)} & 83.2 (7.4\%) & \textbf{\Greencell{88.1 (22.2\%)}} & 86.4 (8.6\%) & 83.9 (15.7\%) \\

&  &\logitsloss & 96.2 (0.4) & \greencell{73.7 (29.8\%)} & \greencell{84.5 (17.5\%)} & 89.0 (11.7\%) & 83.1 (6.6\%) & 87.2 (13.1\%) & 86.5 (10.5\%) & \greencell{83.5 (15.8\%)} \\

&  & \repreloss & 95.9 (0.1) & \greencell{71.9 (24.1\%)} & 83.1 (7.3\%) & 89.2 (14.3\%) & 83.2 (7.4\%) & 86.5 (6.1\%) & 86.3 (9.2\%) & 82.8 (11.8\%) \\

\cline{2-3}
& DeleteWord & \augloss & 95.9 (0.1) & 65.4 (3.5\%) & 83.2 (8.0\%) & 89.3 (15.6\%) & 83.2 (7.4\%) & 87.5 (16.2\%) & 86.7 (13.0\%) & 81.7 (10.1\%) \\

\cline{2-3}
& SubWord & \augloss &96.1 (0.3) & 65.9 (5.1\%) & 83.5 (10.2\%) & 89.3 (15.6\%) & 82.7 (3.7\%) & 86.8 (9.1\%) & 86.3 (9.5\%) & 81.6 (8.7\%) \\

\cline{2-3}
& InsertWord & \augloss &95.8 (0.0) & 66.3 (6.3\%) & \redcell{81.8 (-2.2\%)} & 88.2 (1.3\%) & \redcell{81.9 (-2.2\%)} & 86.2 (3.0\%) & 85.4 (0.7\%) & 80.9 (1.3\%) \\

\cline{2-3}
& \multirow{3}{*}{SpeechAug} & \augloss & 96.0 (0.2) & 70.5 (19.7\%) & 83.7 (11.7\%) & 89.3 (15.6\%) & 82.9 (5.1\%) &\greencell{87.7 (18.2\%)}  & 86.7 (13.0\%) & 82.8 (14.1\%) \\

&  & \logitsloss &96.2 (0.4) & 68.8 (14.3\%) & 82.2 (0.7\%) & 88.4 (3.9\%) & 83.1 (6.6\%) & \greencell{87.7 (18.2\%)} & 86.4 (9.6\%) & 82.0 (8.7\%) \\

&  & \repreloss &96.1 (0.3) & 66.2 (6.0\%) & 82.8 (5.1\%) & 88.5 (5.2\%) & 82.9 (5.1\%) & 86.3 (4.0\%) & 85.9 (4.8\%) & 81.3 (5.1\%) \\

\hline
\multirow{12}{*}{\makecell{Feature-level}} & \multirow{3}{*}{Adv} & \augloss
&96.7 (0.9) & 67.9 (11.4\%) & 84.0 (13.9\%) & \textbf{\Greencell{90.6 (32.5\%)}} & \greencell{84.5 (16.9\%)} & 87.6 (17.2\%) & \textbf{\Greencell{87.6 (22.2\%)}} & \greencell{82.9 (18.4\%)} \\

& & \logitsloss &96.3 (0.5) & 67.7 (10.8\%) & 83.8 (12.4\%) & 89.8 (22.1\%) & 83.7 (11.0\%) & 86.8 (9.1\%) & \greencell{86.8 (14.1\%)} & 82.4 (13.1\%) \\

& & \repreloss & 96.4 (0.6) & 66.6 (7.3\%) & 82.5 (2.9\%) & 89.2 (14.3\%) & 84.0 (13.2\%) & 86.4 (5.1\%) & 86.5 (10.9\%) & 81.7 (8.6\%) \\

\cline{2-3}
& \multirow{3}{*}{TokenCut} & \augloss & 96.5 (0.7) & 70.9 (21.0\%) & \textbf{\Greencell{84.7 (19.0\%)}} & \greencell{90.1 (26.0\%)} & \textbf{\Greencell{84.5 (16.9\%)}} & 87.3 (14.1\%) & \greencell{87.3 (19.0\%)} & \textbf{\Greencell{83.5 (19.4\%)}} \\

& & \logitsloss &96.3 (0.5) & 68.5 (13.3\%) & \greencell{84.6 (18.2\%)} & 89.2 (14.3\%) & 83.1 (6.6\%) & 86.5 (6.1\%) & 86.3 (9.0\%) & 82.4 (11.7\%) \\

& & \repreloss & 96.1 (0.3) & 66.1 (5.7\%) & 82.5 (2.9\%) & 89.4 (16.9\%) & 83.3 (8.1\%) & 87.0 (11.1\%) & 86.6 (12.0\%) & 81.6 (8.9\%) \\

\cline{2-3}
& \multirow{3}{*}{FeatureCut} & \augloss & 95.9 (0.1) & 66.7 (7.6\%) & 83.0 (6.6\%) & 89.9 (23.4\%) & 83.3 (8.1\%) & 86.9 (10.1\%) & 86.7 (13.9\%) & 82.0 (11.2\%) \\

& & \logitsloss &  96.1 (0.3) & 66.3 (6.3\%) & 82.7 (4.4\%) & 89.4 (16.9\%) & 83.0 (5.9\%) & 87.0 (11.1\%) & 86.5 (11.3\%) & 81.7 (8.9\%) \\

& & \repreloss & 96.3 (0.5) & 67.8 (11.1\%) & 82.7 (4.4\%) & 89.1 (13.0\%) & 83.5 (9.6\%) & 86.6 (7.1\%) & 86.4 (9.9\%) & 81.9 (9.0\%) \\

\cline{2-3}
& \multirow{3}{*}{Dropout} & \augloss & 96.0 (0.2) & 66.4 (6.7\%) & 82.4 (2.2\%) & \greencell{90.0 (24.7\%)} & 82.9 (5.1\%) & 86.9 (10.1\%) & 86.6 (13.3\%) & 81.7 (9.8\%) \\

& & \logitsloss & 95.9 (0.1) & 65.7 (4.4\%) & 82.9 (5.8\%) & 89.3 (15.6\%) & 83.5 (9.6\%) & 86.6 (7.1\%) & 86.5 (10.7\%) & 81.6 (8.5\%) \\

& & \repreloss & 96.6 (0.8) & 67.9 (11.4\%) & 83.2 (8.0\%) & 89.2 (14.3\%) &\greencell{84.2 (14.7\%)}  & 86.5 (6.1\%) & 86.7 (11.7\%) & 82.2 (10.9\%) \\

\bottomrule[1pt]
\end{tabular}
    \vspace{-0.2cm}
    \caption{The performance ($F1$ score) of the LSTM-based model with different augmentation strategies on Noise-SF. For cells in  \textit{Baseline} row and \textit{Clean test} column, the numbers in the parenthesis indicate the change of $F1$ score over the baseline (95.8), while for other cells, the numbers in the parenthesis indicate the relative denoising rate (\ro). In \textit{Overall} column, we calculate the average $F1$ and $\rho$ of the five noises respectively. The highest $\rho$ in each column are marked dark green, the second and third highest $\rho$ are marked light green, and the negative $\rho$ are marked red.
    }
    \label{tab:main table}
    \vspace{-0.5cm}
\end{table*}

%

\vspace{-0.2cm}
\subsection{Settings}
\vspace{-0.2cm}
There are two model settings: Glove-Bi-LSTM and BERT-Bi-LSTM. We apply Glove-6B-300d and char embedding\footnote{Char embedding is introduced to tackle out-of-vocabulary.} and BERT-large-uncased as the embedding layer, respectively. We train and evaluate models on the proposed dataset Noise-SF\footnote{We focus on the robustness evaluation, so we don't distinguish the domains.}. All experimental results are repeated three times with different random seeds under the same settings and the average results are reported. \footnote{The experiment details are in the Appendix \ref{appendixd}.}

\vspace{-0.2cm}

\subsection{Comprehensive Comparison Results}

In this section, we employ Bi-LSTM as the backbone and explore the robustness performance of augmentation methods combined with three consistency training strategies\footnote{As DeleteWord, SubWord and InsertWord would change the length of tokens, so the logits consistency and representation consistency are inapplicable.}. Experimental results are illustrated in Table \ref{tab:main table}. Next, we will analyze the effects of various robust training methods from four perspectives: noise level, augmentation level, augmentation type, and training strategy.


\vspace{-0.1cm}

\paragraph{Noise Level}
Comparing the results in the column \ding{172} with columns \ding{173} to \ding{177}, we find that fine-grained noise has a greater effect on baseline performance (e.g., character-level (-31.5), sentence-level (-7.7~-13.6)). Since the original model relies too much on the tokens of slot entity mentions, and fine-grained noise would break slot entity tokens more severely, which leads to the original model perform poorly. In addition, the denoising effect of CharAug and TokenCut on Typos is better than other methods in text-level and feature-level, respectively. When we apply ChauAug to perform random character transformation or TokenCut to randomly mask some token features, the model would be forced to focus on the contextual semantics rather than the slot entity mentions themselves. 
Besides, we also note that most robust training methods are not effective to sentence-level noise. As sentence-level noise is derived from different expression habits in human language, it is hard to simulate it by simple rule-based augmentation methods. 

\vspace{-0.1cm}

\begin{figure}[t]
\centering

\subfigure{
\resizebox{.45\textwidth}{!}{\includegraphics{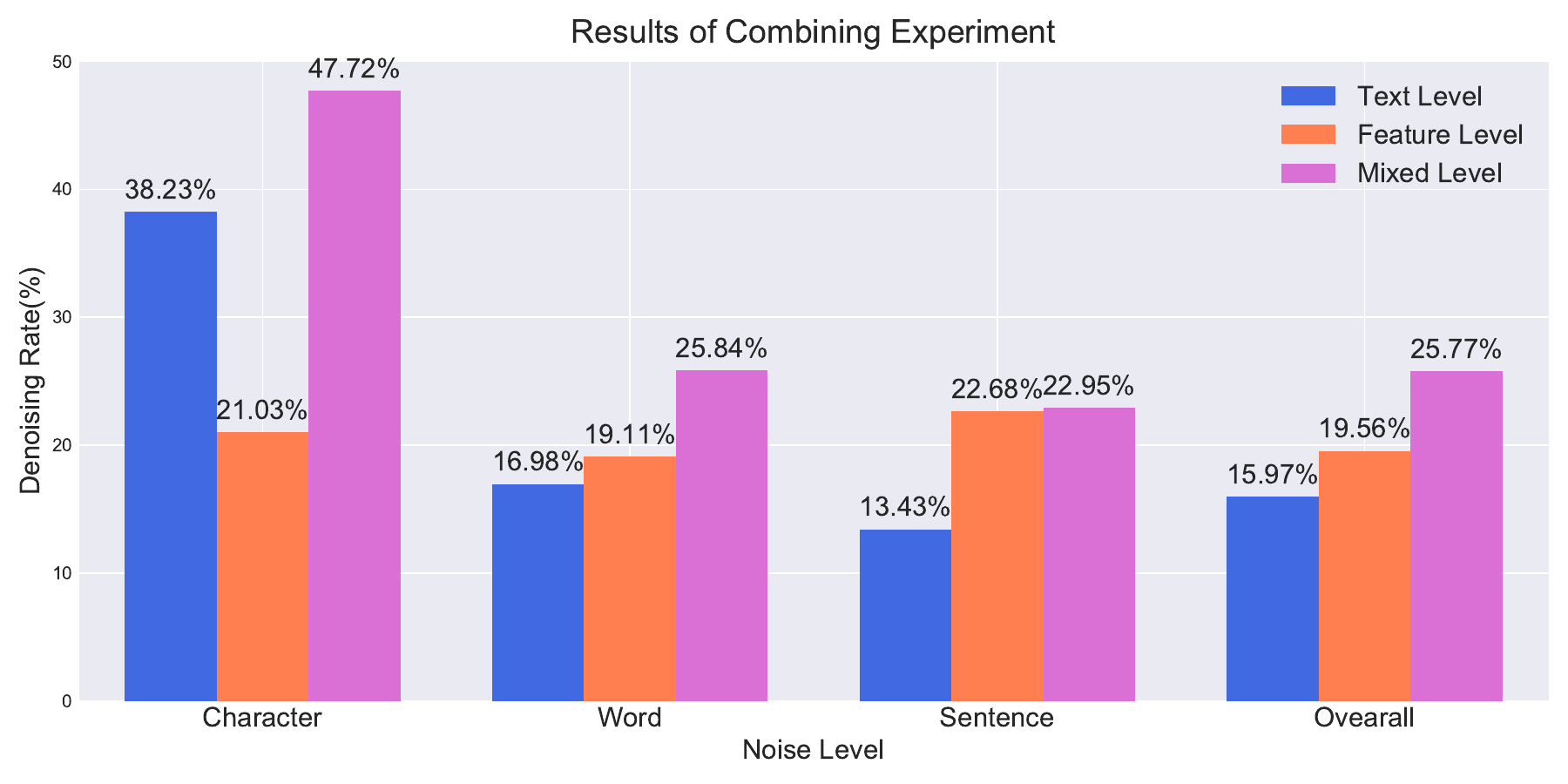}}
}
\vspace{-0.5cm}
\caption{The performance of text-level, feature-level and mixed-level augmentation strategies combined with character, word and sentence-level of noise, respectively.}
\vspace{-0.7cm}
\label{fig:comb2}
\end{figure}

\paragraph{Augmentation Level}
Comparing the results of text-level and feature-level, we find that text-level methods and feature-level methods have the highest relative denoising rate at character-level and sentence-level noise,  respectively (these phenomena can be observed in Figure \ref{fig:comb2}). As discussed above, although CharAug can better reduce the dependence of the model on slot entity mentions, it is difficult to simulate sentence-level noise, which is more complex. In fact, the text-level augmentation method is a discrete interpolation method, which has a discontinuous effect on the contextual semantics in the utterances in the semantic space. On the contrary, the feature-level augmentation method is a continuous interpolation method that can simulate the semantic shift of utterance caused by sentence-level noise at a finer granularity. Therefore, the feature-level augmentation methods are more applicable for sentence-level input noises.


\paragraph{Augmentation Type}
In the previous section, we have analyzed the reasons why CharAug, Adv, and TokenCut performed well on different types of noise. Meanwhile, we find that some other methods did not achieve the denoising effect we expected. For example, DeleteWord does not perform well on Simplification, and InsertWord even has a negative impact on Verbosity. These results suggest that it is very difficult to simulate the complex and diverse sentence-level noise by artificially designed rules. In addition, we observe that in feature-level augmentation, FeatureCut and Dropout, which also belong to the same randomly feature dropping as TokenCut, perform worse than TokenCut significantly. 
A possible explanation for this might be explored by analyzing the distinction of these methods.
In fact, FeatureCut and Dropout would randomly destroy the representation of the entire sequence, while affecting both the semantics of slot entity mentions and context, while tokenCut only performs feature dropping on partial tokens.
Therefore, the augmented samples generated by the former two augmentation methods have less effect on model robustness.
\vspace{-0.2cm}

\paragraph{Training Strategies}

Comparing the three consistency strategies, we find that the effect of augmentation loss is superior to the other two strategies, demonstrating that it is simple and effective to regard augmented samples as extended training data. While training the models, we found that the value of the other two losses are much smaller than cross-entropy loss, and are also affected by different data augmentation methods. Therefore, performance might be better if the weights of the two Losses are carefully adjusted\footnote{For ease of comparison, the weights of all consistency loss are set to 1.0.}.

\vspace{-0.2cm}
\paragraph{Conclusions}
According to the above experimental phenomena and analyses, we can summarize the impact of input noise on the slot filling model into two perspectives: 1) corruption of slot entity mentions; 2) corruption of contextual semantics. Character-level noise mainly break the slot entity mentions, while sentence-level noise (Paraphrase, Verbosity, Simplification) mainly changes the context. The performance of various noise-robust methods shows that: 1) We can reduce the dependence of the slot filling model on slot entity mentions by character-level augmentation methods; 2) To deal with the corruption of contextual semantics, feature-level augmentation (continuous interpolation) methods will be more effective.

\begin{figure}[t]
\centering

\resizebox{.40\textwidth}{!}{\includegraphics{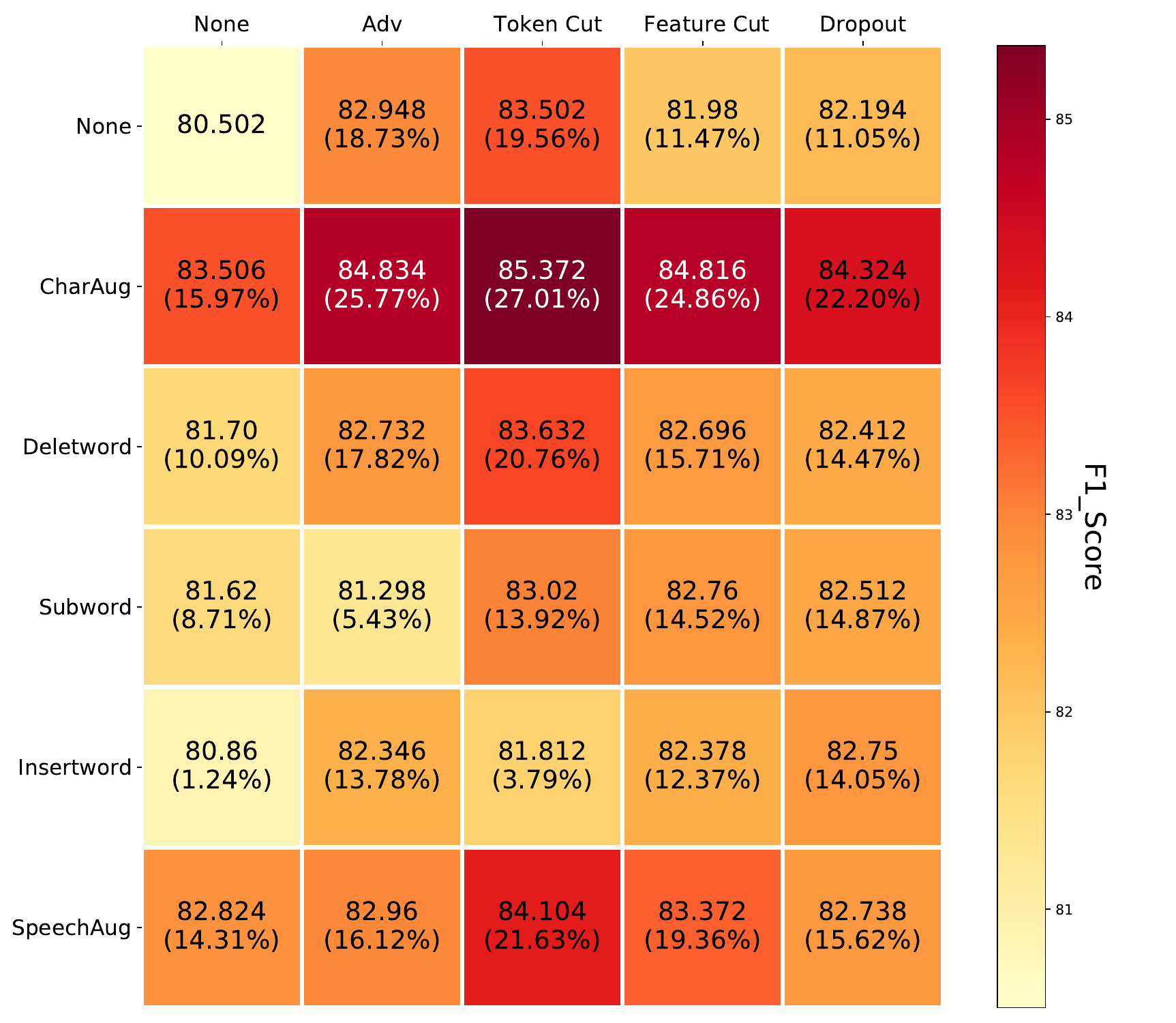}}
\vspace{-0.3cm}
\caption{Visualization of the performance with different
combinations of data augmentation strategies in two levels. The row
indicates the text-level augmentation strategies, while the
column indicates the feature-level augmentation strategies.}
\vspace{-0.3cm}
\label{fig:comb1}
\end{figure}



\begin{table*}[htbp]
\centering
\tiny
\setlength{\tabcolsep}{3.8pt}
\renewcommand{\arraystretch}{1.5}
\begin{tabular}{c| c| c| c| c| c| c c c c| c}
\toprule

\multirow{2}{*}{\textbf{Encoder type}} & \multirow{2}{*}{\textbf{Augmentation type}} & \multirow{2}{*}{\textbf{Loss type}}
& \multirow{2}{*}{\textbf{Clean test}} & \textbf{Character-level}  &  \textbf{Word-level} 
& \multicolumn{4}{c}{\textbf{Sentence-level}} \vline & \multirow{2}{*}{\textbf{Overall}} \\

\cline{5-10}
 & & & & \textbf{Typos} & \textbf{Speech} & \textbf{Paraphrase} 
& \textbf{Verbosity} & \textbf{Simplification} & \textbf{Overall} & \\

\hline

Baseline & none & $\mathcal{L}$
& \underline{96.2} & \makecell{\underline{67.6 (-28.6)}} & \makecell{\underline{82.8 (-13.4)}} & \makecell{\underline{90.4 (-5.8)}} 
& \makecell{\underline{84.4 (-11.8)}} & \makecell{\underline{87.7 (-8.5)}} & \underline{87.5 (-8.7)} & \makecell{\underline{82.6 (-13.6)}} \\

\hline
\multirow{5}{*}{\makecell{Text-level}} & CharAug (NAT) & \augloss & 96.0 (-0.2) & \textcolor[RGB]{0,119,51}{\textbf{78.0 (36.4\%)} } & 85.0 (16.4\%) & 89.9 (-8.6\%) & 83.8 (-5.1\%) & 86.9 (-9.4\%) & 86.9 (-7.7\%) & \textcolor[RGB]{0,119,51}{\textbf{84.7 (5.9\%)}} \\

\cline{2-3}
& DeleteWord & \augloss & 96.0 (-0.2) & 68.5 (3.1\%) & 84.5 (12.7\%) & 90.0 (-6.9\%) & 84.5 (0.8\%) & 88.0 (3.5\%) & 87.5 (-0.8\%) & 83.1 (2.7\%) \\

\cline{2-3}
& SubWord & \augloss & 96.3 (0.1) & 69.6 (7.0\%) & 84.1 (9.7\%) & 90.2 (-3.4\%) & 84.1 (-2.5\%) & 88.2 (5.9\%) & 87.5 (0.0\%) & 83.2 (3.3\%) \\

\cline{2-3}
& InsertWord & \augloss & 96.3 (0.1) & 69.6 (7.0\%) & 84.1 (9.7\%) & 90.2 (-3.4\%) & 84.1 (-2.5\%) & 88.2 (5.9\%) & 87.5 (0.0\%) & 83.2 (3.3\%) \\

\cline{2-3}
& SpeechAug & \augloss  & 95.8 (-0.4) & 72.3 (16.4\%) & \textcolor[RGB]{0,119,51}{\textbf{85.8 (22.4\%})} & 90.2 (-3.4\%) & 82.4 (-16.9\%) & 87.5 (-2.4\%) & 86.7 (-7.6\%) & 83.6 (3.2\%) \\

\hline
\multirow{4}{*}{\makecell{Feature-level}} & Adv & \augloss & 96.4 (0.2) & 68.9 (4.5\%) & 83.8 (7.5\%) & 90.3 (-1.7\%) & 84.5 (0.8\%) & 87.9 (2.4\%) & 87.6 (0.5\%) & 83.1 (2.7\%) \\

\cline{2-3}
& TokenCut & \repreloss & 96.2 (0.0) & 70.9 (11.5\%) & 83.5 (5.2\%) & 90.5 (1.7\%) & 84.2 (-1.7\%) & \textcolor[RGB]{0,119,51}{\textbf{88.8 (12.9\%)}} & 87.8 (4.3\%) & 83.6 (5.9\%) \\

\cline{2-3}
& FeatureCut & \repreloss & 96.2 (0.0) & 68.7 (3.8\%) & 83.8 (7.5\%) & 90.4 (0.0\%) & 
\textcolor[RGB]{0,119,51}{\textbf{84.8 (3.4\%)}} & 87.9 (2.4\%) & 87.7 (1.9\%) & 83.1 (3.4\%) \\

\cline{2-3}
& Dropout & \repreloss & 96.4 (0.2) & 68.8 (4.2\%) & 83.5 (5.2\%) & \textcolor[RGB]{0,119,51}{\textbf{90.8 (6.9\%)}} & 84.3 (-0.8\%) & 88.5 (9.4\%) & \textcolor[RGB]{0,119,51}{\textbf{87.9 (5.2\%)}} & 83.2 (5.0\%) \\

\bottomrule
\end{tabular}
\vspace{-0.3cm}
\caption{The performance of BERT-based model with different augmentation strategies on clean and noisy test sets.}
\label{tab:bert table}
\vspace{-0.3cm}
\end{table*}


\begin{table*}[htbp]
    \centering
    \tiny
    \renewcommand{\arraystretch}{1.6}
    \setlength{\tabcolsep}{3pt}

    \begin{tabular}{c|cccccccc}
        \hline
        \multirow{2}{*}{\textbf{Method}}& \textbf{Char} & \textbf{Word} & \textbf{Sen} & \textbf{Char + Word} & \textbf{Word + Sen} & \textbf{Char + Sen} & \textbf{Char + Word + Sen} & \multirow{2}{*}{\textbf{Overall}} \\
        
        \cline{2-8}
         & \textbf{Typos} & \textbf{SSWN} & \textbf{APP} & \textbf{Typos + SSWN} & \textbf{SSWN + App} & \textbf{Typos + App} & \textbf{Typos + SSWN + APP} & \\
        
        \hline
        Baseline (LSTM) &\underline{55.7 (-38.2)}  & \underline{80.8 (-13.1)} & \underline{71.2 (-22.7)} & \underline{47.5 (-46.4)} & \underline{60.5 (-33.4)} & \underline{38.5 (-55.4)} & \underline{33.0 (-60.9)} &  \underline{55.3 (-38.6)} \\
        
        \hline
        CharAug (NAT) + Adv (\augloss) & 75.9 (52.9\%) & 87.0 (47.3\%) & 82.1 (48.0\%) & 68.9 (46.1\%) & 75.6 (45.2\%) & 63.8 (45.7\%) & 56.6 (38.8\%) & 73.5 (46.3\%) \\

        \hline
        CharAug (NAT) + TokenCut (\augloss) & 75.8 (52.6\%) & 85.9 (38.9\%) & 76.9 (25.1\%) & 69.2 (46.8\%) & 71.3 (32.3\%) & 59.6 (38.1\%) & 52.9 (32.7\%) & 70.2 (38.0\%) \\
        
        \hline

    \end{tabular}
    \vspace{-0.2cm}
    \caption{The performance of two best mixed augmentation strategies (e.g., ``CharAug + Adv'' with Augmentation Loss) on mixed multiple noise SNIPS dataset. SSWN denotes SwapSynWordNet, APP stands for Appendirr and Sen stands for Sentence-level.}
    \label{tab:snips mix}
    \vspace{-0.3cm}
\end{table*}

\subsection{Denoising on Contextual Embedding}

We test our proposed augmentation methods on the model which employs BERT as embedder, and experimental results are shown in Table \ref{tab:bert table}. It is found that: (1) Similar to LSTM, Typos has the most impact on BERT. The impact of other noises is similar to that of the embedder model based on word vectors; (2) Even though the robustness training methods have a good relative denoising effect against Typos, it is still far from being robust to Typos; (3) It is surprising that our proposed methods have no effect on the sentence-level noise and even have a negative effect. It is a chance that as BERT is a context-based embedder, when facing utterances with sentence-level noise, the variation of noisy sentences in the hidden layer is more obvious than that of embedder with word vector. Augmentation and consistency training will increase the instability of internal representation and thus increase the negative impact of noise on the model.


\vspace{-0.1cm}

\subsection{Combining Experiment}
In this section, we first explore the noise robustness for a combination of augmentation methods from different levels (text-level and feature-level)\footnote{Combining experiments of the same level (text-level \& text-level, feature-level \& feature-level) are in the Appendix \ref{appendixe}}. 

The results are shown in Figure \ref{fig:comb1}. 
First, CharAug and TokenCut are the most effective augmentation methods at the text-level and feature-level, respectively. When the two methods are combined, the noise robustness of LSTM is further improved and reaches the best (27.01\%). Secondly, when we take the text-level method as the benchmark, the introduction of the feature-level method basically achieves a stable improvement, which shows that the feature-level method is a more generalized method that will bring improvement if jointly used with other methods.

In addition, as shown in Figure \ref{fig:comb2}, we compare the two-level methods and their combinations horizontally at three noise levels. We find that the combination of the two methods can significantly further improve the denoising performance of the model. However, on the sentence-level noise, the denoising performance of the mixed level is basically the same as that of the feature-level only. This phenomenon shows that it does not make much sense to supplement the coarse-grained discrete interpolation method to the fine-grained continuous interpolation method, which further implicates that to solve the overall semantic shift of sentences caused by sentence-level noise, it is necessary to explore robust representation methods from the perspective of semantic features.
\vspace{-0.2cm}

\subsection{SNIPS multi-noise Experiment}
\vspace{-0.1cm}
As in real dialogue scenarios, mixed multiple noises often appear in an input utterance at the same time, which is more complex. In this subsection, we explore whether noise-robust training has good performance in mixed noise scenarios.

\vspace{-0.2cm}
\paragraph{Settings}
Based on SNIPS \cite{coucke2018snips}, we utilize textflint \cite{gui2021textflint} to introduce character-level noise Typos, Word-level noise SwapSynWordNet, sentence-level noise appendirr, and mixed multiple noise to the test set and construct a multi-noise evaluation set.
\vspace{-0.3cm}

\paragraph{Results \& Analysis}
As shown in Table \ref{tab:snips mix}, both entity mentions and contextual semantics are corrupted in mixed multiple noise scenarios, resulting in a catastrophic drop in model performance. We find that the combination of text-level and feature-level methods can effectively improve the noise robustness of the model. The overall relative denoising rate of the CharAug + Adv method  is close to 50\%. The model maintains an almost 40\% relative denoising rate even with the joint disturbances from three-level noises, which shows the effectiveness and stability of our methods.
In addition, we also explore the performance of various denoising methods alone on this mixed multiple noise dataset. The experimental results are in Appendix \ref{appendixf}.
\vspace{-0.2cm}

\section{Discussion \& Future Work}
\vspace{-0.2cm}

Recent studies \cite{lin-etal-2020-rigorous,10.1162/coli_a_00397,NERcase2020} show that, the performance of sequence labeling models affected by the tokens of entity mentions and contexts, and the main factor driving high performance is fitting the entity tokens themselves. In our experimental analysis, we also divide the impact of input noise into two perspectives. On the one hand, the input noise corrupts the slot entity mentions, and on the other hand, the input noise makes the contextual semantics of the slot entity shifted. Considering these two perspectives, we find that character-level noise mainly affects slot entity mentions, while sentence-level noise mainly affects contextual semantics\footnote{We refer to Appendix \ref{appendixg} for statistical results.}. Comparing the denoising performance of various data augmentation methods, we find that methods such as \emph{CharAug} have better denoising performance on character-level noise, and feature-level augmentation (continuous interpolation) methods have better denoising performance on sentence-level noise. In addition, simulation multi-noise experiments further verify our conclusions.


Considering the above discussion, in future work: (1) We can explore the robustness of the slot filling model from the perspectives of reducing the model's dependence on slot entity mentions and improving the diversity of contextual semantics; 
(2) By simulating the corruption of slot entity mentions, the slot filling model can be forced to reduce its dependence on slot entity mentions; 
(3) We should explore robust training methods from high-dimensional contextual semantic features to improve the robustness of the model to contextual semantic shifts; 
(4) More complex noise scenarios need to be explored (e.g., real mixed multi-noise, inconsistently distributed noise in multi-domain dialogues and so on.


\vspace{-0.2cm}

\section{Conclusion}
\vspace{-0.2cm}
In this paper, we explore the effects of different robust-training strategies toward various input noise that exist in real dialogue scenarios. A crowd-sourced real noise-evaluation dataset is introduced, and exhaustive experiments and insightful analyses are provided to fuel research in this direction. Finally, we make some forward-looking suggestions for further research.

\section*{Limitations}
As we explored in experiments, the input noise has a serious impact on slot filling model, and the researches in this direction is greatly hindered by the lack of noise data of real dialogue scenarios. Although this paper introduces a human-annotated multi-type noise dataset and explores the effect of various methods on different noise, there are still some limitations in our works. In general, it can be summarized as the following three aspects: 1) the methods proposed in this paper is mainly for the purposes of exploration and therefore lacks a deeper investigation of noise-robust training; 2) the dataset in this paper is also crowdsourced due to the difficulty of collecting noise in real dialogue scenarios;\footnote{We discuss this topic in detail in Appendix \ref{appendixc}} 3) the noise in real scenarios is often mixed, multi-domain and even multi-modal, but these noise types in the complex scenes are not addressed in this paper.




\section*{Ethics Statement}
\subsection*{1. Broader Impact}

%
In recent years, supervised deep neural network models have demonstrated remarkable performance in many natural language processing tasks and thus have been applied to a wide range of real-world applications. In particular, automatic dialogue systems have been widely used in various realistic conversation scenarios. Slot filling models as an essential component of the dialogue system, which plays a critical role in processing utterances of users. It is obvious there are various noises (text perturbations) that exist in realistic user's utterances input which would limit the performance of slot filling models. However, the existing slot filling methods pay little attention to this problem or only study the input noise with the rule-based constructed dataset, which 
prevents the development of noise-robust slot filling methods. This work introduces a human-annotated multi-type noise dataset for evaluating slot filling in a more realistic conversation scenario. We also propose a series of robust training methods to explore the noise-robust slot filling models. Nevertheless, there are still many limitations in our work, as we discussed above. Therefore, we call on more researchers to pay attention to this issue and make more efforts to promote the application of NLP models in practical scenarios.

\subsection*{2. Privacy and Licensing Issues}
Noise-SF is constructed from the public dataset RADDLE \cite{peng2020raddle}. As their paper mentions, all language variations corpus is created by workers on Amazon Mechanical Turks. Moreover, all the information such as phone numbers and reference codes in the dialogues is fictitious. Therefore, our Noise-SF does not involve any privacy-related issues.
\subsection*{3. Annotator Information and Compensation}
We recruit five graduate students who are proficient in English to carry out the annotating work of the nature of volunteer activities. After completing the annotating work, each annotator can get one credit in the name of volunteer activity, which is very important to meet the graduation requirements. We add the compensation rules into the recruitment document and all the annotators accept it as a reasonable standard before they agree to annotate the data. 
The five annotators are all Chinese college students aged 20-23, whose mother tongue is Chinese, including 3 males and 2 females. 



\bibliography{anthology,custom}

\begin{thebibliography}{58}
\expandafter\ifx\csname natexlab\endcsname\relax\def\natexlab#1{#1}\fi

\bibitem[{Agarwal et~al.(2021)Agarwal, Yang, Wallace, and Nenkova}]{10.1162/coli_a_00397}
Oshin Agarwal, Yinfei Yang, Byron~C. Wallace, and Ani Nenkova. 2021.
\newblock \href {https://doi.org/10.1162/coli_a_00397} {{Interpretability Analysis for Named Entity Recognition to Understand System Predictions and How They Can Improve}}.
\newblock \emph{Computational Linguistics}, 47(1):117--140.

\bibitem[{Belinkov and Bisk(2017)}]{belinkov2017synthetic}
Yonatan Belinkov and Yonatan Bisk. 2017.
\newblock Synthetic and natural noise both break neural machine translation.
\newblock \emph{arXiv preprint arXiv:1711.02173}.

\bibitem[{Brill and Moore(2000)}]{brill2000improved}
Eric Brill and Robert~C Moore. 2000.
\newblock An improved error model for noisy channel spelling correction.
\newblock In \emph{Proceedings of the 38th annual meeting of the association for computational linguistics}, pages 286--293.

\bibitem[{Budzianowski et~al.(2020)Budzianowski, Wen, Tseng, Casanueva, Ultes, Ramadan, and Gašić}]{budzianowski2020multiwoz}
Paweł Budzianowski, Tsung-Hsien Wen, Bo-Hsiang Tseng, Iñigo Casanueva, Stefan Ultes, Osman Ramadan, and Milica Gašić. 2020.
\newblock \href {http://arxiv.org/abs/1810.00278} {Multiwoz -- a large-scale multi-domain wizard-of-oz dataset for task-oriented dialogue modelling}.

\bibitem[{Chakraborty et~al.(2018)Chakraborty, Alam, Dey, Chattopadhyay, and Mukhopadhyay}]{chakraborty2018adversarial}
Anirban Chakraborty, Manaar Alam, Vishal Dey, Anupam Chattopadhyay, and Debdeep Mukhopadhyay. 2018.
\newblock \href {http://arxiv.org/abs/1810.00069} {Adversarial attacks and defences: A survey}.

\bibitem[{Chiu and Nichols(2016)}]{chiu2016named}
Jason~PC Chiu and Eric Nichols. 2016.
\newblock Named entity recognition with bidirectional lstm-cnns.
\newblock \emph{Transactions of the Association for Computational Linguistics}, 4:357--370.

\bibitem[{Coucke et~al.(2018)Coucke, Saade, Ball, Bluche, Caulier, Leroy, Doumouro, Gisselbrecht, Caltagirone, Lavril, Primet, and Dureau}]{coucke2018snips}
Alice Coucke, Alaa Saade, Adrien Ball, Théodore Bluche, Alexandre Caulier, David Leroy, Clément Doumouro, Thibault Gisselbrecht, Francesco Caltagirone, Thibaut Lavril, Maël Primet, and Joseph Dureau. 2018.
\newblock \href {http://arxiv.org/abs/1805.10190} {Snips voice platform: an embedded spoken language understanding system for private-by-design voice interfaces}.

\bibitem[{Coulombe(2018)}]{coulombe2018text}
Claude Coulombe. 2018.
\newblock \href {http://arxiv.org/abs/1812.04718} {Text data augmentation made simple by leveraging nlp cloud apis}.

\bibitem[{Dong et~al.(2022)Dong, Guo, Wang, Li, Wang, Zeng, He, Zhao, Lei, Cui et~al.}]{dong2022pssat}
Guanting Dong, Daichi Guo, Liwen Wang, Xuefeng Li, Zechen Wang, Chen Zeng, Keqing He, Jinzheng Zhao, Hao Lei, Xinyue Cui, et~al. 2022.
\newblock Pssat: A perturbed semantic structure awareness transferring method for perturbation-robust slot filling.
\newblock \emph{arXiv preprint arXiv:2208.11508}.

\bibitem[{Dong et~al.(2023{\natexlab{a}})Dong, Li, Wang, Zhang, Xian, and Xu}]{dong2023bridging}
Guanting Dong, Rumei Li, Sirui Wang, Yupeng Zhang, Yunsen Xian, and Weiran Xu. 2023{\natexlab{a}}.
\newblock Bridging the kb-text gap: Leveraging structured knowledge-aware pre-training for kbqa.
\newblock \emph{arXiv preprint arXiv:2308.14436}.

\bibitem[{Dong et~al.(2023{\natexlab{b}})Dong, Wang, Wang, Guo, Fu, Wu, Zeng, Li, Hui, He, Cui, Gao, and Xu}]{10095149}
Guanting Dong, Zechen Wang, Liwen Wang, Daichi Guo, Dayuan Fu, Yuxiang Wu, Chen Zeng, Xuefeng Li, Tingfeng Hui, Keqing He, Xinyue Cui, Qixiang Gao, and Weiran Xu. 2023{\natexlab{b}}.
\newblock \href {https://doi.org/10.1109/ICASSP49357.2023.10095149} {A prototypical semantic decoupling method via joint contrastive learning for few-shot named entity recognition}.
\newblock In \emph{ICASSP 2023 - 2023 IEEE International Conference on Acoustics, Speech and Signal Processing (ICASSP)}, pages 1--5.

\bibitem[{Dong et~al.(2023{\natexlab{c}})Dong, Wang, Zhao, Zhao, Guo, Fu, Hui, Zeng, He, Li et~al.}]{dong2023multi}
Guanting Dong, Zechen Wang, Jinxu Zhao, Gang Zhao, Daichi Guo, Dayuan Fu, Tingfeng Hui, Chen Zeng, Keqing He, Xuefeng Li, et~al. 2023{\natexlab{c}}.
\newblock A multi-task semantic decomposition framework with task-specific pre-training for few-shot ner.
\newblock \emph{arXiv preprint arXiv:2308.14533}.

\bibitem[{Fang et~al.(2020)Fang, Filice, Limsopatham, and Rokhlenko}]{fang2020using}
Anjie Fang, Simone Filice, Nut Limsopatham, and Oleg Rokhlenko. 2020.
\newblock Using phoneme representations to build predictive models robust to asr errors.
\newblock In \emph{Proceedings of the 43rd International ACM SIGIR Conference on Research and Development in Information Retrieval}, pages 699--708.

\bibitem[{Fu et~al.(2020)Fu, Liu, and Zhang}]{NERcase2020}
Jinlan Fu, Pengfei Liu, and Qi~Zhang. 2020.
\newblock \href {https://doi.org/10.1609/aaai.v34i05.6276} {Rethinking generalization of neural models: A named entity recognition case study}.
\newblock \emph{Proceedings of the AAAI Conference on Artificial Intelligence}, 34(05):7732–7739.

\bibitem[{Gardner et~al.(2018)Gardner, Neumann, Grus, and Lourie}]{gardner-etal-2018-writing}
Matt Gardner, Mark Neumann, Joel Grus, and Nicholas Lourie. 2018.
\newblock \href {https://aclanthology.org/D18-3003} {Writing code for {NLP} research}.
\newblock In \emph{Proceedings of the 2018 Conference on Empirical Methods in Natural Language Processing: Tutorial Abstracts}, Melbourne, Australia. Association for Computational Linguistics.

\bibitem[{Goo et~al.(2018)Goo, Gao, Hsu, Huo, Chen, Hsu, and Chen}]{goo-etal-2018-slot}
Chih-Wen Goo, Guang Gao, Yun-Kai Hsu, Chih-Li Huo, Tsung-Chieh Chen, Keng-Wei Hsu, and Yun-Nung Chen. 2018.
\newblock \href {https://doi.org/10.18653/v1/N18-2118} {Slot-gated modeling for joint slot filling and intent prediction}.
\newblock In \emph{Proceedings of the 2018 Conference of the North {A}merican Chapter of the Association for Computational Linguistics: Human Language Technologies, Volume 2 (Short Papers)}, pages 753--757, New Orleans, Louisiana. Association for Computational Linguistics.

\bibitem[{Goodfellow et~al.(2014)Goodfellow, Shlens, and Szegedy}]{goodfellow2014explaining}
Ian~J Goodfellow, Jonathon Shlens, and Christian Szegedy. 2014.
\newblock Explaining and harnessing adversarial examples.
\newblock \emph{arXiv preprint arXiv:1412.6572}.

\bibitem[{Gopalakrishnan et~al.(2020)Gopalakrishnan, Hedayatnia, Wang, Liu, and Hakkani-Tur}]{gopalakrishnan2020neural}
Karthik Gopalakrishnan, Behnam Hedayatnia, Longshaokan Wang, Yang Liu, and Dilek Hakkani-Tur. 2020.
\newblock Are neural open-domain dialog systems robust to speech recognition errors in the dialog history? an empirical study.
\newblock \emph{arXiv preprint arXiv:2008.07683}.

\bibitem[{Gui et~al.(2021)Gui, Wang, Zhang, Liu, Zou, Zhou, Zheng, Zhang, Wu, Ye, Pang, Zhang, Li, Ma, Fei, Cai, Zhao, Hu, Yan, Tan, Hu, Bian, Liu, Zhu, Qin, Xing, Fu, Zhang, Peng, Zheng, Zhou, Wei, Qiu, and Huang}]{gui2021textflint}
Tao Gui, Xiao Wang, Qi~Zhang, Qin Liu, Yicheng Zou, Xin Zhou, Rui Zheng, Chong Zhang, Qinzhuo Wu, Jiacheng Ye, Zexiong Pang, Yongxin Zhang, Zhengyan Li, Ruotian Ma, Zichu Fei, Ruijian Cai, Jun Zhao, Xingwu Hu, Zhiheng Yan, Yiding Tan, Yuan Hu, Qiyuan Bian, Zhihua Liu, Bolin Zhu, Shan Qin, Xiaoyu Xing, Jinlan Fu, Yue Zhang, Minlong Peng, Xiaoqing Zheng, Yaqian Zhou, Zhongyu Wei, Xipeng Qiu, and Xuanjing Huang. 2021.
\newblock \href {http://arxiv.org/abs/2103.11441} {Textflint: Unified multilingual robustness evaluation toolkit for natural language processing}.

\bibitem[{Guo et~al.(2023)Guo, Dong, Fu, Wu, Zeng, Hui, Wang, Li, Wang, He, Cui, and Xu}]{10094766}
Daichi Guo, Guanting Dong, Dayuan Fu, Yuxiang Wu, Chen Zeng, Tingfeng Hui, Liwen Wang, Xuefeng Li, Zechen Wang, Keqing He, Xinyue Cui, and Weiran Xu. 2023.
\newblock \href {https://doi.org/10.1109/ICASSP49357.2023.10094766} {Revisit out-of-vocabulary problem for slot filling: A unified contrastive framework with multi-level data augmentations}.
\newblock In \emph{ICASSP 2023 - 2023 IEEE International Conference on Acoustics, Speech and Signal Processing (ICASSP)}, pages 1--5.

\bibitem[{He et~al.(2020{\natexlab{a}})He, Lei, Yang, Jiang, and Wang}]{he-etal-2020-syntactic}
Keqing He, Shuyu Lei, Yushu Yang, Huixing Jiang, and Zhongyuan Wang. 2020{\natexlab{a}}.
\newblock \href {https://doi.org/10.18653/v1/2020.coling-main.246} {Syntactic graph convolutional network for spoken language understanding}.
\newblock In \emph{Proceedings of the 28th International Conference on Computational Linguistics}, pages 2728--2738, Barcelona, Spain (Online). International Committee on Computational Linguistics.

\bibitem[{He et~al.(2020{\natexlab{b}})He, Yan, and Xu}]{he-etal-2020-learning-tag}
Keqing He, Yuanmeng Yan, and Weiran Xu. 2020{\natexlab{b}}.
\newblock \href {https://doi.org/10.18653/v1/2020.acl-main.58} {Learning to tag {OOV} tokens by integrating contextual representation and background knowledge}.
\newblock In \emph{Proceedings of the 58th Annual Meeting of the Association for Computational Linguistics}, pages 619--624, Online. Association for Computational Linguistics.

\bibitem[{Heigold et~al.(2018)Heigold, Varanasi, Neumann, and van Genabith}]{heigold-etal-2018-robust}
Georg Heigold, Stalin Varanasi, G{\"u}nter Neumann, and Josef van Genabith. 2018.
\newblock \href {https://aclanthology.org/W18-1807} {How robust are character-based word embeddings in tagging and {MT} against wrod scramlbing or randdm nouse?}
\newblock In \emph{Proceedings of the 13th Conference of the Association for Machine Translation in the {A}mericas (Volume 1: Research Track)}, pages 68--80, Boston, MA. Association for Machine Translation in the Americas.

\bibitem[{Hinton et~al.(2012)Hinton, Srivastava, Krizhevsky, Sutskever, and Salakhutdinov}]{hinton2012improving}
Geoffrey~E Hinton, Nitish Srivastava, Alex Krizhevsky, Ilya Sutskever, and Ruslan~R Salakhutdinov. 2012.
\newblock Improving neural networks by preventing co-adaptation of feature detectors.
\newblock \emph{arXiv preprint arXiv:1207.0580}.

\bibitem[{Huang and Chen(2020)}]{huang2020learning}
Chao-Wei Huang and Yun-Nung Chen. 2020.
\newblock Learning asr-robust contextualized embeddings for spoken language understanding.
\newblock In \emph{ICASSP 2020-2020 IEEE International Conference on Acoustics, Speech and Signal Processing (ICASSP)}, pages 8009--8013. IEEE.

\bibitem[{Lample et~al.(2016)Lample, Ballesteros, Subramanian, Kawakami, and Dyer}]{lample2016neural}
Guillaume Lample, Miguel Ballesteros, Sandeep Subramanian, Kazuya Kawakami, and Chris Dyer. 2016.
\newblock Neural architectures for named entity recognition.
\newblock \emph{arXiv preprint arXiv:1603.01360}.

\bibitem[{Lei et~al.(2023{\natexlab{a}})Lei, Dong, Wang, Wang, and Wang}]{lei2023instructerc}
Shanglin Lei, Guanting Dong, Xiaoping Wang, Keheng Wang, and Sirui Wang. 2023{\natexlab{a}}.
\newblock \href {http://arxiv.org/abs/2309.11911} {Instructerc: Reforming emotion recognition in conversation with a retrieval multi-task llms framework}.

\bibitem[{Lei et~al.(2023{\natexlab{b}})Lei, Wang, Dong, Li, and Liu}]{lei2023watch}
Shanglin Lei, Xiaoping Wang, Guanting Dong, Jiang Li, and Yingjian Liu. 2023{\natexlab{b}}.
\newblock Watch the speakers: A hybrid continuous attribution network for emotion recognition in conversation with emotion disentanglement.
\newblock \emph{arXiv preprint arXiv:2309.09799}.

\bibitem[{Li et~al.(2020{\natexlab{a}})Li, Liu, Ruan, Soldaini, Hamza, and Su}]{li2020multi}
Mingda Li, Xinyue Liu, Weitong Ruan, Luca Soldaini, Wael Hamza, and Chengwei Su. 2020{\natexlab{a}}.
\newblock Multi-task learning of spoken language understanding by integrating n-best hypotheses with hierarchical attention.
\newblock In \emph{Proceedings of the 28th International Conference on Computational Linguistics: Industry Track}, pages 113--123.

\bibitem[{Li et~al.(2020{\natexlab{b}})Li, Ruan, Liu, Soldaini, Hamza, and Su}]{li2020improving}
Mingda Li, Weitong Ruan, Xinyue Liu, Luca Soldaini, Wael Hamza, and Chengwei Su. 2020{\natexlab{b}}.
\newblock Improving spoken language understanding by exploiting asr n-best hypotheses.
\newblock \emph{arXiv preprint arXiv:2001.05284}.

\bibitem[{Li et~al.(2022)Li, Lei, Wang, Dong, Zhao, Liu, Xu, and Zhang}]{9747192}
Xuefeng Li, Hao Lei, Liwen Wang, Guanting Dong, Jinzheng Zhao, Jiachi Liu, Weiran Xu, and Chunyun Zhang. 2022.
\newblock \href {https://doi.org/10.1109/ICASSP43922.2022.9747192} {A robust contrastive alignment method for multi-domain text classification}.
\newblock In \emph{ICASSP 2022 - 2022 IEEE International Conference on Acoustics, Speech and Signal Processing (ICASSP)}, pages 7827--7831.

\bibitem[{Li et~al.(2023)Li, Wang, Dong, He, Zhao, Lei, Liu, and Xu}]{li-etal-2023-generative}
Xuefeng Li, Liwen Wang, Guanting Dong, Keqing He, Jinzheng Zhao, Hao Lei, Jiachi Liu, and Weiran Xu. 2023.
\newblock \href {https://doi.org/10.18653/v1/2023.findings-acl.52} {Generative zero-shot prompt learning for cross-domain slot filling with inverse prompting}.
\newblock In \emph{Findings of the Association for Computational Linguistics: ACL 2023}, pages 825--834, Toronto, Canada. Association for Computational Linguistics.

\bibitem[{Lin et~al.(2020)Lin, Lu, Tang, Han, Sun, Wei, and Yuan}]{lin-etal-2020-rigorous}
Hongyu Lin, Yaojie Lu, Jialong Tang, Xianpei Han, Le~Sun, Zhicheng Wei, and Nicholas~Jing Yuan. 2020.
\newblock \href {https://doi.org/10.18653/v1/2020.emnlp-main.592} {A rigorous study on named entity recognition: Can fine-tuning pretrained model lead to the promised land?}
\newblock In \emph{Proceedings of the 2020 Conference on Empirical Methods in Natural Language Processing (EMNLP)}, pages 7291--7300, Online. Association for Computational Linguistics.

\bibitem[{Liu and Lane(2016{\natexlab{a}})}]{liu-lane-2016-joint}
Bing Liu and Ian Lane. 2016{\natexlab{a}}.
\newblock \href {https://doi.org/10.18653/v1/W16-3603} {Joint online spoken language understanding and language modeling with recurrent neural networks}.
\newblock In \emph{Proceedings of the 17th Annual Meeting of the Special Interest Group on Discourse and Dialogue}, pages 22--30, Los Angeles. Association for Computational Linguistics.

\bibitem[{Liu and Lane(2016{\natexlab{b}})}]{Liu2016AttentionBasedRN}
Bing Liu and Ian~R. Lane. 2016{\natexlab{b}}.
\newblock Attention-based recurrent neural network models for joint intent detection and slot filling.
\newblock In \emph{INTERSPEECH}.

\bibitem[{Liu et~al.(2020)Liu, Takanobu, Wen, Wan, Li, Nie, Li, Peng, and Huang}]{liu2020robustness}
Jiexi Liu, Ryuichi Takanobu, Jiaxin Wen, Dazhen Wan, Hongguang Li, Weiran Nie, Cheng Li, Wei Peng, and Minlie Huang. 2020.
\newblock Robustness testing of language understanding in task-oriented dialog.
\newblock \emph{arXiv preprint arXiv:2012.15262}.

\bibitem[{Lu et~al.(2021)Lu, Han, Yuan, Wang, Lei, Jiang, and Wu}]{lu-etal-2021-slot}
Hengtong Lu, Zhuoxin Han, Caixia Yuan, Xiaojie Wang, Shuyu Lei, Huixing Jiang, and Wei Wu. 2021.
\newblock \href {https://doi.org/10.18653/v1/2021.findings-acl.440} {Slot transferability for cross-domain slot filling}.
\newblock In \emph{Findings of the Association for Computational Linguistics: ACL-IJCNLP 2021}, pages 4970--4979, Online. Association for Computational Linguistics.

\bibitem[{Miller(1995)}]{miller1995wordnet}
George~A Miller. 1995.
\newblock Wordnet: a lexical database for english.
\newblock \emph{Communications of the ACM}, 38(11):39--41.

\bibitem[{Moradi and Samwald(2021{\natexlab{a}})}]{moradi2021evaluating}
Milad Moradi and Matthias Samwald. 2021{\natexlab{a}}.
\newblock Evaluating the robustness of neural language models to input perturbations.
\newblock \emph{arXiv preprint arXiv:2108.12237}.

\bibitem[{Moradi and Samwald(2021{\natexlab{b}})}]{eval2021}
Milad Moradi and Matthias Samwald. 2021{\natexlab{b}}.
\newblock \href {https://doi.org/10.18653/v1/2021.emnlp-main.117} {Evaluating the robustness of neural language models to input perturbations}.
\newblock \emph{Proceedings of the 2021 Conference on Empirical Methods in Natural Language Processing}.

\bibitem[{Namysl et~al.(2020{\natexlab{a}})Namysl, Behnke, and K{\"o}hler}]{namysl2020nat}
Marcin Namysl, Sven Behnke, and Joachim K{\"o}hler. 2020{\natexlab{a}}.
\newblock Nat: noise-aware training for robust neural sequence labeling.
\newblock \emph{arXiv preprint arXiv:2005.07162}.

\bibitem[{Namysl et~al.(2020{\natexlab{b}})Namysl, Behnke, and K{\"o}hler}]{namysl-etal-2020-nat}
Marcin Namysl, Sven Behnke, and Joachim K{\"o}hler. 2020{\natexlab{b}}.
\newblock \href {https://doi.org/10.18653/v1/2020.acl-main.138} {{NAT}: Noise-aware training for robust neural sequence labeling}.
\newblock In \emph{Proceedings of the 58th Annual Meeting of the Association for Computational Linguistics}, pages 1501--1517, Online. Association for Computational Linguistics.

\bibitem[{Namysl et~al.(2021)Namysl, Behnke, and Köhler}]{nat2021}
Marcin Namysl, Sven Behnke, and Joachim Köhler. 2021.
\newblock \href {https://doi.org/10.18653/v1/2021.findings-acl.27} {Empirical error modeling improves robustness of noisy neural sequence labeling}.
\newblock \emph{Findings of the Association for Computational Linguistics: ACL-IJCNLP 2021}.

\bibitem[{Paszke et~al.(2019)Paszke, Gross, Massa, Lerer, Bradbury, Chanan, Killeen, Lin, Gimelshein, Antiga et~al.}]{paszke2019pytorch}
Adam Paszke, Sam Gross, Francisco Massa, Adam Lerer, James Bradbury, Gregory Chanan, Trevor Killeen, Zeming Lin, Natalia Gimelshein, Luca Antiga, et~al. 2019.
\newblock Pytorch: An imperative style, high-performance deep learning library.
\newblock \emph{Advances in neural information processing systems}, 32:8026--8037.

\bibitem[{Peng et~al.(2020{\natexlab{a}})Peng, Li, Zhang, Zhu, Li, and Gao}]{peng2020raddle}
Baolin Peng, Chunyuan Li, Zhu Zhang, Chenguang Zhu, Jinchao Li, and Jianfeng Gao. 2020{\natexlab{a}}.
\newblock Raddle: An evaluation benchmark and analysis platform for robust task-oriented dialog systems.
\newblock \emph{arXiv preprint arXiv:2012.14666}.

\bibitem[{Peng et~al.(2020{\natexlab{b}})Peng, Zhu, Zeng, and Gao}]{peng2020data}
Baolin Peng, Chenguang Zhu, Michael Zeng, and Jianfeng Gao. 2020{\natexlab{b}}.
\newblock Data augmentation for spoken language understanding via pretrained models.
\newblock \emph{arXiv e-prints}, pages arXiv--2004.

\bibitem[{Ribeiro et~al.(2020)Ribeiro, Wu, Guestrin, and Singh}]{check2020}
Marco~Tulio Ribeiro, Tongshuang Wu, Carlos Guestrin, and Sameer Singh. 2020.
\newblock \href {https://doi.org/10.18653/v1/2020.acl-main.442} {Beyond accuracy: Behavioral testing of nlp models with checklist}.
\newblock \emph{Proceedings of the 58th Annual Meeting of the Association for Computational Linguistics}.

\bibitem[{Rozsa et~al.(2016)Rozsa, Rudd, and Boult}]{rozsa2016adversarial}
Andras Rozsa, Ethan~M Rudd, and Terrance~E Boult. 2016.
\newblock Adversarial diversity and hard positive generation.
\newblock In \emph{Proceedings of the IEEE Conference on Computer Vision and Pattern Recognition Workshops}, pages 25--32.

\bibitem[{Ruan et~al.(2020)Ruan, Nechaev, Chen, Su, and Kiss}]{ruan2020towards}
Weitong Ruan, Yaroslav Nechaev, Luoxin Chen, Chengwei Su, and Imre Kiss. 2020.
\newblock Towards an asr error robust spoken language understanding system.
\newblock In \emph{INTERSPEECH}, pages 901--905.

\bibitem[{Shen et~al.(2020)Shen, Zheng, Shen, Qu, and Chen}]{shen2020simple}
Dinghan Shen, Mingzhi Zheng, Yelong Shen, Yanru Qu, and Weizhu Chen. 2020.
\newblock \href {http://arxiv.org/abs/2009.13818} {A simple but tough-to-beat data augmentation approach for natural language understanding and generation}.

\bibitem[{Sun et~al.(2023)Sun, Gao, Mou, Dong, Liu, and Guo}]{10193387}
Mingyang Sun, Qixiang Gao, Yutao Mou, Guanting Dong, Ruifang Liu, and Wenbin Guo. 2023.
\newblock \href {https://doi.org/10.1109/ICASSPW59220.2023.10193387} {Improving few-shot performance of dst model through multitask to better serve language-impaired people}.
\newblock In \emph{2023 IEEE International Conference on Acoustics, Speech, and Signal Processing Workshops (ICASSPW)}, pages 1--5.

\bibitem[{Wang et~al.(2021)Wang, Li, Liu, He, Yan, and Xu}]{wang-etal-2021-bridge}
Liwen Wang, Xuefeng Li, Jiachi Liu, Keqing He, Yuanmeng Yan, and Weiran Xu. 2021.
\newblock \href {https://aclanthology.org/2021.emnlp-main.746} {Bridge to target domain by prototypical contrastive learning and label confusion: Re-explore zero-shot learning for slot filling}.
\newblock In \emph{Proceedings of the 2021 Conference on Empirical Methods in Natural Language Processing}, pages 9474--9480, Online and Punta Cana, Dominican Republic. Association for Computational Linguistics.

\bibitem[{Wei and Zou(2019)}]{wei2019eda}
Jason Wei and Kai Zou. 2019.
\newblock Eda: Easy data augmentation techniques for boosting performance on text classification tasks.
\newblock \emph{arXiv preprint arXiv:1901.11196}.

\bibitem[{Wu et~al.(2021)Wu, Chen, Ding, and Tao}]{wu2021bridging}
Di~Wu, Yiren Chen, Liang Ding, and Dacheng Tao. 2021.
\newblock Bridging the gap between clean data training and real-world inference for spoken language understanding.
\newblock \emph{arXiv preprint arXiv:2104.06393}.

\bibitem[{Yan et~al.(2020)Yan, He, Xu, Liu, Meng, Hu, and Xu}]{yan-etal-2020-adversarial}
Yuanmeng Yan, Keqing He, Hong Xu, Sihong Liu, Fanyu Meng, Min Hu, and Weiran Xu. 2020.
\newblock \href {https://doi.org/10.18653/v1/2020.emnlp-main.490} {Adversarial semantic decoupling for recognizing open-vocabulary slots}.
\newblock In \emph{Proceedings of the 2020 Conference on Empirical Methods in Natural Language Processing (EMNLP)}, pages 6070--6075, Online. Association for Computational Linguistics.

\bibitem[{Yan et~al.(2021)Yan, Li, Wang, Zhang, Wu, and Xu}]{yan2021consert}
Yuanmeng Yan, Rumei Li, Sirui Wang, Fuzheng Zhang, Wei Wu, and Weiran Xu. 2021.
\newblock \href {http://arxiv.org/abs/2105.11741} {Consert: A contrastive framework for self-supervised sentence representation transfer}.

\bibitem[{Zeng et~al.(2022)Zeng, He, Wang, Fu, Dong, Geng, Wang, Wang, Sun, Wu et~al.}]{zeng2022semi}
Weihao Zeng, Keqing He, Zechen Wang, Dayuan Fu, Guanting Dong, Ruotong Geng, Pei Wang, Jingang Wang, Chaobo Sun, Wei Wu, et~al. 2022.
\newblock Semi-supervised knowledge-grounded pre-training for task-oriented dialog systems.
\newblock \emph{arXiv preprint arXiv:2210.08873}.

\bibitem[{Zhao et~al.(2022)Zhao, Dong, Shi, Yan, Xu, and Li}]{zhao-etal-2022-entity}
Gang Zhao, Guanting Dong, Yidong Shi, Haolong Yan, Weiran Xu, and Si~Li. 2022.
\newblock \href {https://doi.org/10.18653/v1/2022.findings-emnlp.473} {Entity-level interaction via heterogeneous graph for multimodal named entity recognition}.
\newblock In \emph{Findings of the Association for Computational Linguistics: EMNLP 2022}, pages 6345--6350, Abu Dhabi, United Arab Emirates. Association for Computational Linguistics.

\end{thebibliography}
\renewcommand\thesection{\Alph{section}}
\appendix
 
\section{Data Processing}
\label{appendixa}

We directly extract slot information from RADDLE as a clean test set. We also do manual annotations carefully on user utterances with five types of noise as the noise test benchmark set of Noise-SF. Note that, there are one-to-one correspondences between noise test data and clean test data. The annotation team is composed of five well-trained human annotators, and each annotator independently annotates five types of noise data. Each type is 800 utterances with a total of 4000. Then we summarize and filter the independent annotation results, finally obtaining a high-quality noisy dataset with 98\% (manual evaluation) annotation accuracy. To our knowledge, Noise-SF is the first real-world multi-noise evaluation dataset for slot filling.

\section{Statistic of Noise-SF}
\label{appendixb}

\begin{table}[ht]
    \centering
    \tiny
    \setlength{\tabcolsep}{1.5pt}
    \renewcommand{\arraystretch}{1.5}
    \begin{tabular}{c|ccccc}
        \hline
        \multirow{2}{*}{\textbf{↓Domain}} & \multicolumn{5}{c}{\textbf{Noise-SF (Typos / Speech Errors / Simplification / 
        Paraphrase / Verbosity)}} \\
        
        \cline{2-6}
        & \textbf{Clean train} & \textbf{Clean valid} & \textbf{Clean test} & \textbf{Noise Evaluation} & \textbf{Slot types} \\
        
        \hline
        Attraction & 12023 & 1073 & 74 & 74 $\times$ 5 & 3\\
        
        \hline
        Hotel & 15035 & 1339 & 224 & 224 $\times$ 5 & 3\\
        
        \hline
        Restaurant & 16034 & 1428 & 256 & 256 $\times$ 5 & 6\\
        
        \hline
        Train & 13027 & 1160 & 307 & 307 $\times$ 5 & 7\\
        
        \hline
        Total & 56119 & 5000 & 861 & 861 $\times$ 5 & 14\\
        
        \hline
    \end{tabular}
    \vspace{-0.2cm}
    \caption{The statistics of Noise-SF. For every sample in Clean test, there are corresponding noise-input pairs for five types of noise.}
    \vspace{-0.4cm}
    \label{tab:dataset table}
\end{table}


Noise-SF contains a clean training set, a clean validation set, a clean test set, and an evaluation set containing five types of real dialogue input noise. Noise-SF contains data from four different domains. Consequently, it can also be used for noise test in multi-domain slot filling tasks. The evaluation result on the clean test set is considered as a baseline for comparative analysis of the further noise-robust evaluation. Table \ref{tab:dataset table} shows the statistical information of the noise-input evaluation set and  Table \ref{tab:data samples} demonstrates the examples in 5 different domains and 6 different noises in our Noise-SF. 

\section{The consideration of noise evaluation dataset}
\label{appendixc}

\begin{figure*}[ht]
\centering
\resizebox{.8\textwidth}{!}{\includegraphics{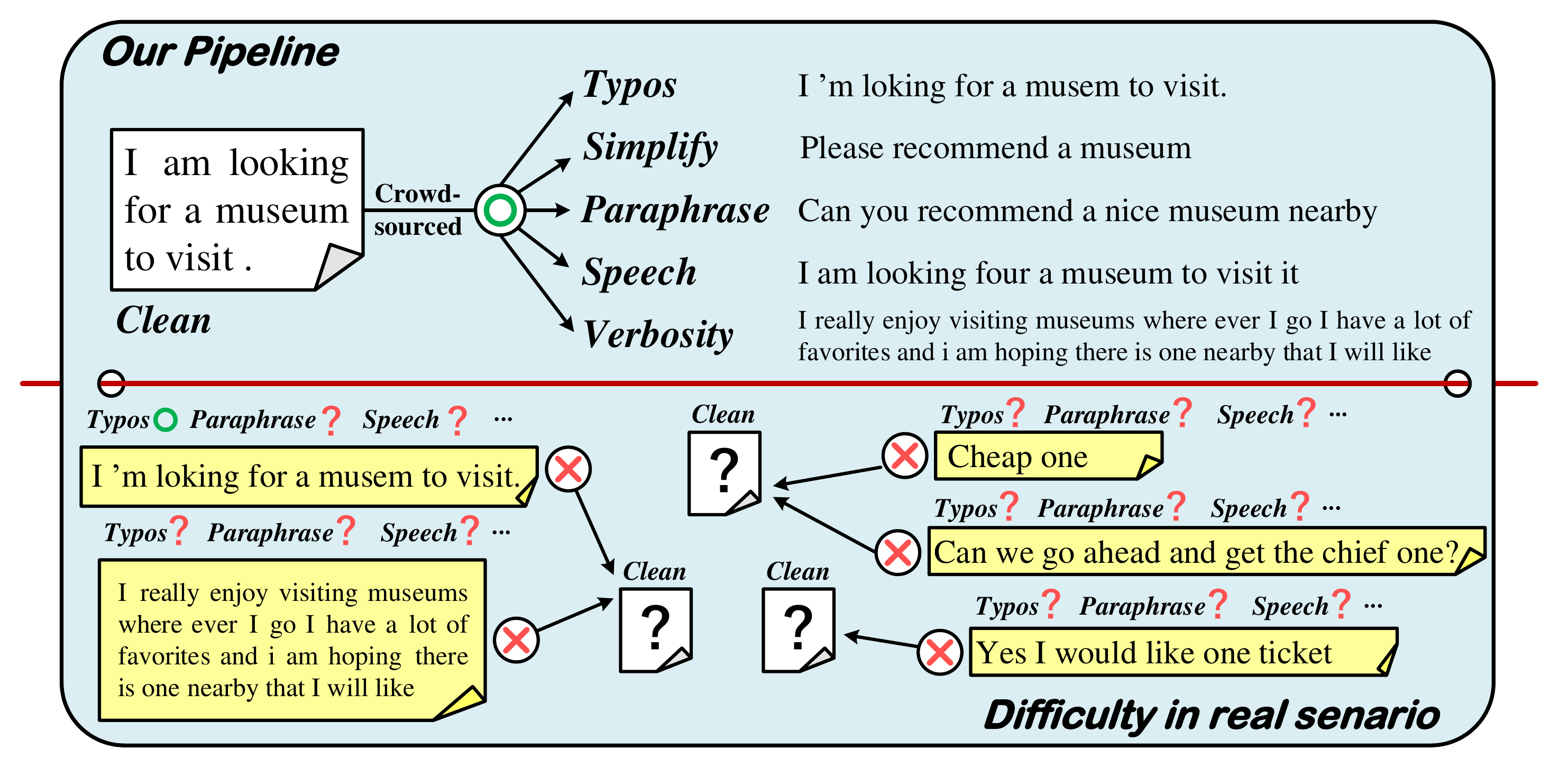}}
\caption{Our pipeline and difficulty of collecting noisy samples in real scenario}
\label{fig:response}
\end{figure*}
In this paper, we aim to construct a dataset containing the original clean training set and a noisy test set with multiple types of input noise corresponding to the original test samples. Based on this benchmark, we explore how various noises affect the performance of the slot filling model, the effectiveness of robust training framework and the effects of different augmentation methods.

Therefore, the key point of our work is training the model on the clean training set, and then comparing the model performance on the clean test set and the various noise evaluation set. 
However, due to the large randomness and bias, there is almost no way to infer the original noise-free utterances and corresponding noise type from the noisy utterances, as shown in the Figure \ref{fig:response}. Besides, noisy texts in real scenarios won't appear in the form of the complete group of all noise types. Therefore, it is difficult to construct complete original-noisy sample groups, like:
\begin{equation}
\begin{aligned}
\vspace{-0.3cm}
\text{clean}\ \text{Text}_A:\ \{\text{Text}_A ^ {\text{noise}_1},\text{Text}_A ^ {\text{noise}_2},\dots \}
\end{aligned}
\end{equation}
These difficulties prevent us from directly constructing the noise evaluation dataset with the noisy texts from real scenarios.
RADDLE \cite{peng2020raddle} offers a crowd-sourced robustness evaluation benchmark for Dialog State Tracking DST, which includes various noisy utterances existed in real dialogue scenarios.
Although there may be some differences between the crowd-sourced noisy utterances and the real noise utterances, the evaluation data in the form of $\text{clean}\_\text{text} : \{\text{noisy}\_\text{texts}\}$ is more convenient and valuable to analyze the essence of the influence of various noises on the slot filling models and explore noise-robust model training methods from different perspectives.

\section{Experiment Details}
\label{appendixd}

Our methods are implemented with PyTorch \cite{paszke2019pytorch} and AllenNLP \cite{gardner-etal-2018-writing}. We take Bi-LSTM as the mainly analyzed model. The hidden size of Bi-LSTM is set to 128 and the dropout rate is set to 0.2. The transform probability $p$ for Character-level is set to 0.15, and the transform probability $p$ for Word-level and Sentence-level is set to 0.3. In the training stage, the weights of consistency loss are set to 1.0. For all the experiments, we train and test our model on the 2080Ti GPU. It takes an average of 1.5 hours to run with 12 epochs on Noise-SF dataset.

\begin{figure}[htbp]

\centering
\resizebox{.4\textwidth}{!}{\includegraphics{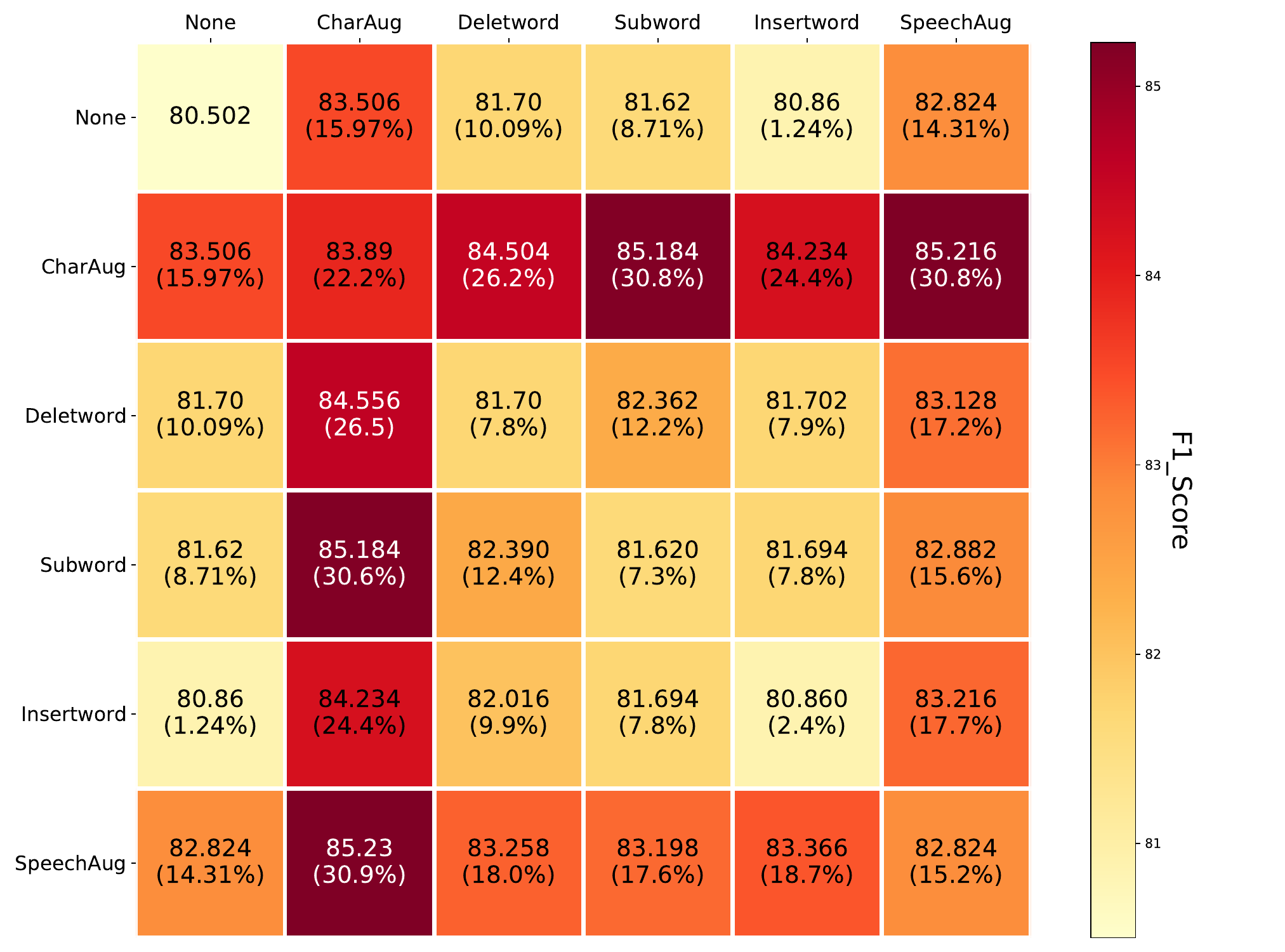}}
\caption{Visualization of the performance with text-level data augmentation strategies. }
\label{fig:heat input}

\end{figure}

\begin{figure}[htbp]

\centering
\resizebox{.4\textwidth}{!}{\includegraphics{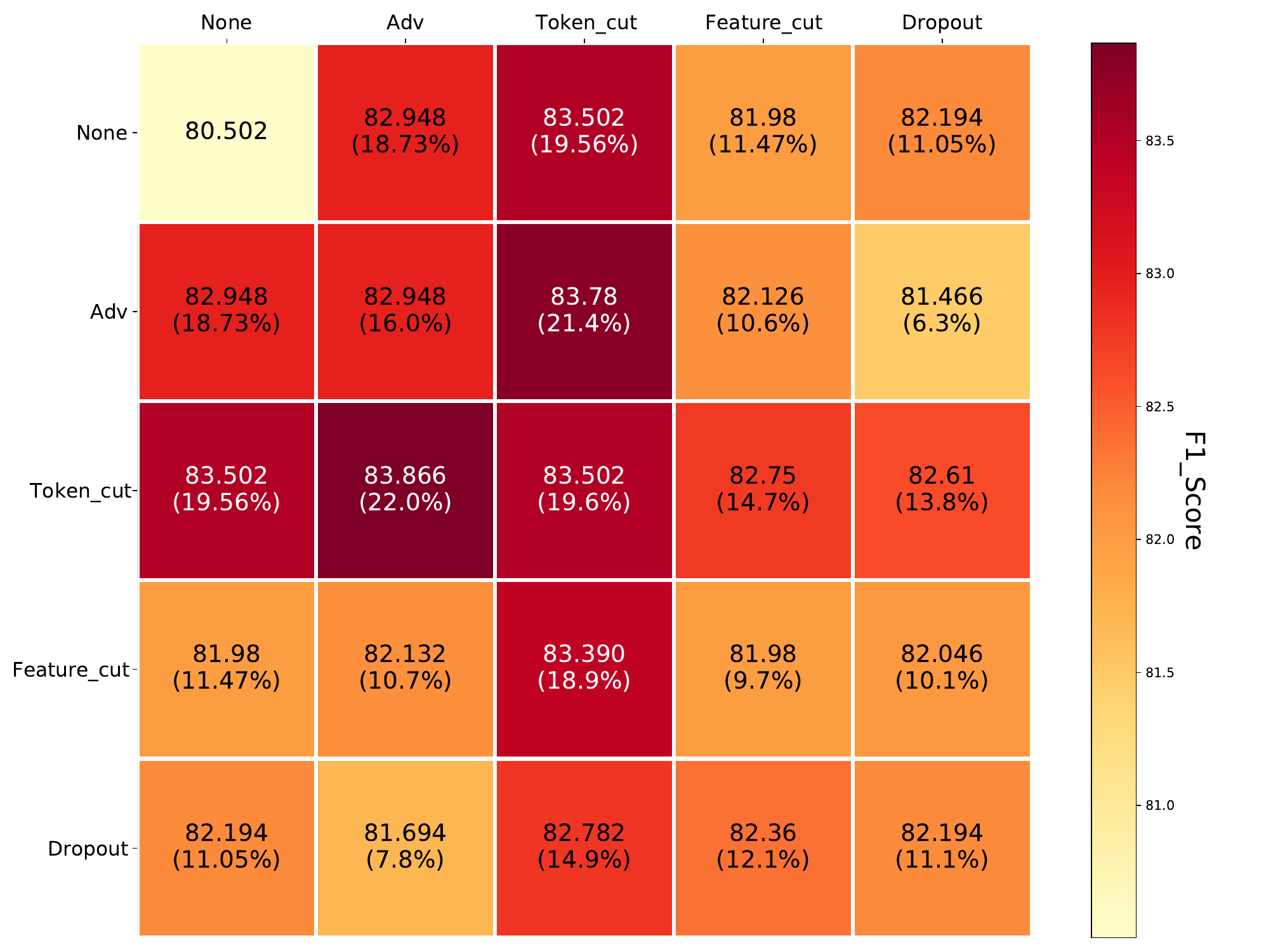}}
\caption{Visualization of the performance with feature-level data augmentation strategies.}
\label{fig:heat feature}
\end{figure}
\section{Combining Experiments of the same level}
\label{appendixe}

 In this section, we explore the noise robustness for a combination of augmentation strategies from the same level (text-level and text-level, feature-level and feature-level), Figure \ref{fig:heat input} and \ref{fig:heat feature} shows the Visualization of the performance with different combinations of data augmentation strategies in text-level and feature-level. It can be found that the relative denoising rate fluctuates greatly with different text-level augmentation strategies. However, feature-level augmentation strategies are more generalized method that will bring steady improvement.

\section{SNIPS noise experiment}
\label{appendixf}

Based on the SNIPS \cite{coucke2018snips}, we introduce 7 different types of noise, which belong to Character-level, Word-level and Sentence-level. The Table \ref{tab:snips results} shows the denoising effect of Text-level and Feature-level augmentation strategies in different noisy data. We can get a conclusion similar to the previous experiment, the fine-grained noise has a greater impact on the performance of the original model and the denoising effect of CharAug and TokenCut on Typos is better than other methods in text-level and feature-level, respectively. Besides, most robust training methods are not effective in denoising the sentence-level noise. These observation motivates us to explore robust training methods from high-dimensional contextual semantic features to improve the robustness of the model.

\section{The statistical results of noise damage}
\label{appendixg}

We make a statistical analysis of the damage type of the input noise to the original data from the perspectives of slot entity mentions and contextual semantics.
To evaluate the contextual semantics damage rate ($D_{CS}$) and slot entity mentions damage rate ($D_{SEM}$), we define them in (3) and (4) :
\begin{equation}
\begin{aligned}
{D_{CS}} = \frac{{N}_{\rm clean}^c-{N}_{\rm noise}^c}{{N}_{\rm clean}^c} 
\end{aligned}
\end{equation}

\begin{equation}
\begin{aligned}
{D_{SEM}} = \frac{{N}_{\rm clean}^s-{N}_{\rm noise}^s}{{N}_{\rm clean}^s} 
\end{aligned}
\end{equation}
, where ${N}_{\rm clean}^s$, ${N}_{\rm clean}^c$ indicate the number of slot entities and non-slot entities in the clean test set, respectively. And ${N}_{\rm noise}^s$, ${N}_{\rm noise}^c$ denote the number of slot entities and non-slot entities which are not changed under specific noise perturbation.
Table \ref{tab:damage} shows the statistical results of noise damage in two datasets, we find that character-level noise mainly affects slot entity mentions, while sentence-level noise mainly affects contextual semantics. The conclusion is similar to the previous experiment analysis.

\begin{table}[htbp]
  \centering
  \tiny
  
    \begin{tabular}{|c|c|c|c|}
    \hline
    Dataset & Noise Type & $D_{CS}$ & $D_{SEM}$ \bigstrut\\
    \hline
    \multirow{5}[10]{*}{Noise-SF} & Typos & 62.6\%  & 38.5\% \bigstrut\\
\cline{2-4}          & Paraphrase & 52.6\% & 19.7\% \bigstrut\\
\cline{2-4}          & Simplification & 67.0\%  & 21.3\% \bigstrut\\
\cline{2-4}          & Speech & 17.4\% & 27.9\% \bigstrut\\
\cline{2-4}          & Verbosity & 39.9\% & 17.8\% \bigstrut\\
    \hline
    \multirow{7}[14]{*}{SNIPS multi-noise} & Typos & 51.7\% & 44.5\% \bigstrut\\
\cline{2-4}          & SSWN   & 57.8\% & 13.7\% \bigstrut\\
\cline{2-4}          & APP & 52.3\% & 0\% \bigstrut\\
\cline{2-4}          & Typos+SSWN & 58.5\% & 53.1\% \bigstrut\\
\cline{2-4}          & App+SSWN & 56.8\% & 11.4\% \bigstrut\\
\cline{2-4}          & Typos+APP & 51.7\% & 44.8\% \bigstrut\\
\cline{2-4}          & Typos+SSWN+APP & 58.5\% & 53.0\% \bigstrut\\
    \hline
    \end{tabular}%
    \caption{The statistical results of noise damage. $D_{CS}$ denotes contextual semantics damage rate, $D_{SEM}$ stands for slot entity mentions damage rate. SSWN and APP represent SwapSynWordNet and Appendirr.}
  \label{tab:damage}%
\end{table}%

        
        
        
        
        
        
        


\vspace{-0.8cm}

        
        
        
        
        
        
        

\def \mc#1{\makecell[c]{#1}}

\begin{table*}[   ]
  \centering
  \tiny
\setlength{\tabcolsep}{2.5pt}
\renewcommand{\arraystretch}{1.5}
    \begin{tabular}{c|p{10em}|p{10em}|p{8em}|p{10em}|p{10em}|p{17em}}
    \toprule
    \multicolumn{1}{p{4.055em}|}{\mc{\textbf{Field}}} & \mc{\textbf{Clean} }& \mc{\textbf{Paraphrase}} & \mc{\textbf{Simplification}} & \mc{\textbf{Speech}} & \mc{\textbf{Typos}} & \mc{\textbf{Verbosity}} \\
    \hline
    \multicolumn{1}{c|}{\multirow{2}[4]{*}{Attraction}} & i am looking for a museum to visit . & can you recommend a nice museum nearby & please recommend a museum & i am looking four a museum two visit it & i 'm loking for a musem to visit . & i really enjoy visiting museums where ever i go i have a lot of favorites and i am hoping there is one nearby that i will like \\
\cline{2-7}          & O O O O O B-type O O O & O O O O O B-type O & O O O B-type & O O O O O B-type O O O & O O O O O B-type O O O & O O O O B-type O O O O O O O O O O O O O O O O O O O O O O \\
    \hline
    \multicolumn{1}{c|}{\multirow{2}[4]{*}{Hotel}} & can we go ahead and get the cheap one please & i want to proceed with the inexpensive place & cheap one & can we go ahead and get the chief one please & cen we go ahead and get teh cheep one ? & i 'm thinking about it and i think that for me economically at least the right decision will have to be the place with a pricing structure on the cheap end \\
\cline{2-7}          & O O O O O O O B-price O O & O O O O O O B-price O & B-price O & O O O O O O O B-price O O & O O O O O O O B-price O O & O O O O O O O O O O O O O O O O O O O O O O O O O O O O O B-price O \\
    \hline
    \multicolumn{1}{c|}{\multirow{2}[4]{*}{Hotel}} & what about moderately priced 4 star lodging of any hotel type  & how about a 4 star moderately priced room of any kind & moderately priced 4 star any hotel type & what about moderately priced four star loging of any hotel tab & whta about mod priced 4star lodging of any hotel tpe & if there are no matches can you locate a 4 star moderately priced lodging of any hotel type \\
\cline{2-7}          & O O B-price O B-stars O O O O O O & O O O B-stars O B-price O O O O O & B-price O B-stars O O O O & O O B-price O B-stars O O O O O O & O O B-price O B-stars O O O O O & O O O O O O O O O B-stars O B-price O O O O O O \\
    \hline
    \multicolumn{1}{c|}{\multirow{2}[4]{*}{Restaurant}} & is there a restaurant that serves indian food in the west part of town ? & i would like information about a indian restaurant in the west part of town & give me an indian restaurant in the west please & is there a restaurant that serves in d and food in the west part of town & is there a restrant thate serves indand fod in the west pat of towne ? & could you please inform me about an indian restaurant in the west part of town since my first option is not available \\
\cline{2-7}          & O O O O O O B-food O O O B-area O O O O & O O O O O O B-food O O O B-area O O O & O O O B-food O O O B-area O & O O O O O O B-food I-food I-food O O O B-area O O O & O O O O O O B-food O O O B-area O O O O & O O O O O O O B-food O O O B-area O O O O O O O O O O \\
    \hline
    \multicolumn{1}{c|}{\multirow{2}[4]{*}{Train}} & yes i would like a ticket for one . & yes i would like one ticket & yes one ticket & yes i would like a ticket for wun & yus i wod lik a tiket for onee . & yes i would appreciate if you could book a ticket for me \\
\cline{2-7}          & O O O O O O O B-people O & O O O O B-people O & O B-people O & O O O O O O O B-people & O O O O O O O B-people O & O O O O O O O O O O O O \\
    \bottomrule
    \end{tabular}%
  \caption{The data samples of Noise-SF dataset.}
  \label{tab:data samples}%
\end{table*}%

\def \u#1{\underline{#1}}
\begin{table*}[htbp]
  \centering
  \tiny
  \setlength{\tabcolsep}{3pt}
  \renewcommand{\arraystretch}{1.3}
    \begin{tabular}{c|c|c|c|cc|cc|cc|c}
    \toprule
    \multirow{2}{*}{\textbf{Encoder type}} & \multirow{2}{*}{\textbf{Augmentation type}} & \multirow{2}{*}{\textbf{Loss type} } & \multirow{2}{*}{\textbf{Clean test}} & \multicolumn{2}{c|}{\textbf{Character-level}} & \multicolumn{2}{c|}{\textbf{Word-level}} & \multicolumn{2}{c|}{\textbf{Sentence-level}} & \multirow{2}{*}{\textbf{Overall}} \\
\cline{5-10}          &       &       &       & \textbf{EntTypos} & \textbf{Keyboard} & \textbf{Asr\_noise} & \textbf{SpellingError} & \textbf{AppendIrr} & \textbf{ConcatSent} &  \\
    \hline
    Baseline & none  & $\mathcal{L}$  & \u{93.9}  & \u{55.7 (-38.2)} & \u{73.1 (-20.8)}& \u{90.3 (-3.6)} & \u{62.2 (-31.7)} & \u{71.2 (-22.7)} & \u{85.0 (-8.9)} & \u{72.9 (-21.0)} \\
    \hline
    \multirow{9}{*}{Text-level} & \multirow{3}{*}{CharAug} & \augloss & 93.7 (-0.2) & 77.7 (57.6\%) & 83.5 (50.0\%) & 90.0 (-8.3\%) & 75.8 (42.9\%) & 74.0 (12.3\%) & 85.2 (2.2\%) & 81.0 (38.6\%) \\
\cline{3-3}          &       & \logitsloss & 76.3 (-17.6) & 62.8 (18.6\%) & 66.4 (-32.2\%) & 70.4 (-552.8\%) & 58.9 (-10.4\%) & 59.6 (-51.1\%) & 53.1 (-358.4\%) & 61.9 (-52.4\%) \\
\cline{3-3}          &       & \repreloss & 91.4 (-2.5) & 63.5 (20.4\%) & 76.8 (17.8\%) & 87.9 (-66.7\%) & 67.2 (15.8\%) & 71.1 (-0.4\%) & 88.2 (36.0\%) & 75.8 (13.8\%) \\
\cline{2-3}          & DeleteWord & \augloss & 93.8 (-0.1) & 55.0 (-1.8\%) & 72.3 (-3.8\%) & 90.4 (2.8\%) & 61.6 (-1.9\%) & 69.2 (-8.8\%) & 85.3 (3.4\%) & 72.3 (-2.9\%) \\
\cline{2-3}          & SubWord & \augloss & 93.8 (-0.1) & 55.8 (0.3\%) & 75.0 (9.1\%) & 91.4 (30.6\%) & 65.7 (11.0\%) & 73.3 (9.3\%) & 84.1 (-10.1\%) & 74.2 (6.2\%) \\
\cline{2-3}          & InsertWord & \augloss & 92.8 (-1.1) & 56.5 (2.1\%) & 75.3 (10.6\%) & 89.8 (-13.9\%) & 63.9 (5.4\%) & 80.5 (41.0\%) & 82.1 (-32.6\%) & 74.7 (8.6\%) \\
\cline{2-3}          & \multirow{3}{*}{SpeechAug} & \augloss & 93.7 (-0.2) & 68.0 (32.2\%) & 79.1 (28.8\%) & 91.4 (30.6\%) & 70.1 (24.9\%) & 72.8 (7.0\%) & 86.4 (15.7\%) & 78.0 (24.3\%) \\
\cline{3-3}          &       & \logitsloss & 85.1 (-8.8) & 58.6 (7.6\%) & 72.6 (-2.4\%) & 81.9 (-233.3\%) & 62.2 (0.0\%) & 70.2 (-4.4\%) & 70.4 (-164.0\%) & 69.3 (-17.1\%) \\
\cline{3-3}          &       & \repreloss & 92.5 (-1.4) & 55.2 (-1.3\%) & 74.1 (4.8\%) & 89.0 (-36.1\%) & 64.5 (7.3\%) & 77.6 (28.2\%) & 89.2 (47.2\%) & 74.9 (9.5\%) \\
    \hline
    \multirow{12}{*}{Feature-level} & \multirow{3}{*}{Adv} & \augloss & 95.4 (1.5) & 57.4 (4.5\%) & 75.2 (10.1\%) & 92.0 (47.2\%) & 63.6 (4.4\%) & 77.0 (25.6\%) & 88.0 (33.7\%) & 75.5 (12.4\%) \\
\cline{3-3}          &       & \logitsloss & 93.9 (0.0) & 54.8 (-2.4\%) & 75.8 (13.0\%) & 91.9 (44.4\%) & 63.3 (3.5\%) & 79.4 (36.1\%) & 86.8 (20.2\%) & 75.3 (11.4\%) \\
\cline{3-3}          &       & \repreloss & 94.8 (0.9) & 53.7 (-5.2\%) & 74.9 (8.7\%) & 91.3 (27.8\%) & 62.5 (0.9\%) & 77.9 (29.5\%) & 88.6 (40.4\%) & 74.8 (9.0\%) \\
\cline{2-3}          & \multirow{3}{*}{TokenCut} & \augloss & 94.8 (0.9) & 65.5 (25.7\%) & 79.8 (32.2\%) & 91.7 (38.9\%) & 67.9 (18.0\%) & 74.0 (12.3\%) & 86.0 (11.2\%) & 77.5 (21.9\%) \\
\cline{3-3}          &       & \logitsloss & 92.4 (-1.5) & 63.3 (19.9\%) & 80.2 (34.1\%) & 90.0 (-8.3\%) & 71.4 (29.0\%) & 80.2 (39.6\%) & 72.4 (-141.6\%) & 76.3 (16.2\%) \\
\cline{3-3}          &       & \repreloss & 91.2 (-2.7) & 54.3 (-3.7\%) & 74.4 (6.3\%) & 88.5 (-50.0\%) & 63.5 (4.1\%) & 70.3 (-4.0\%) & 84.5 (-5.6\%) & 72.6 (-1.4\%) \\
\cline{2-3}          & \multirow{3}{*}{FeatureCut} & \augloss & 94.8 (0.9) & 57.6 (5.0\%) & 73.7 (2.9\%) & 91.2 (25.0\%) & 62.5 (0.9\%) & 73.8 (11.5\%) & 86.7 (19.1\%) & 74.2 (6.2\%) \\
\cline{3-3}          &       & \logitsloss & 93.8 (-0.1) & 55.6 (-0.3\%) & 75.8 (13.0\%) & 91.1 (22.2\%) & 63.0 (2.5\%) & 78.1 (30.4\%) & 81.2 (-42.7\%) & 74.1 (5.7\%) \\
\cline{3-3}          &       & \repreloss & 93.6 (-0.3) & 53.4 (-6.0\%) & 72.9 (-1.0\%) & 89.4 (-25.0\%) & 60.2 (-6.3\%) & 75.3 (18.1\%) & 77.0 (-89.9\%) & 71.4 (-7.1\%) \\
\cline{2-3}          & \multirow{3}{*}{Dropout} & \augloss & 94.5 (0.6) & 56.2 (1.3\%) & 73.3 (1.0\%) & 90.9 (16.7\%) & 61.6 (-1.9\%) & 71.5 (1.3\%) & 86.4 (15.7\%) & 73.3 (1.9\%) \\
\cline{3-3}          &       & \logitsloss & 93.8 (-0.1) & 56.0 (0.8\%) & 77.0 (18.8\%) & 91.5 (33.3\%) & 66.0 (12.0\%) & 77.8 (29.1\%) & 82.5 (-28.1\%) & 75.1 (10.5\%) \\
\cline{3-3}          &       & \repreloss & 93.9 (0.0) & 53.3 (-6.3\%) & 72.5 (-2.9\%) & 89.8 (-13.9\%) & 60.3 (-6.0\%) & 76.5 (23.3\%) & 78.5 (-73.0\%) & 71.8 (-5.2\%) \\
    \bottomrule
    \end{tabular}%
  \caption{Experimental results on the SNIPS dataset.}
  \label{tab:snips results}%
\end{table*}%

\end{document}